\pdfoutput=1

\documentclass[11pt]{article}

\usepackage{acl}

\usepackage{times}
\usepackage{latexsym}

\usepackage[T1]{fontenc}

\usepackage[utf8]{inputenc}

\usepackage{microtype}

\usepackage{inconsolata}

\usepackage{graphicx}

\usepackage{multirow}
\usepackage{url}
\usepackage{booktabs}
\usepackage{linguex}
\usepackage{amsmath} 
\usepackage{subfig}
\usepackage{tabularx}
\usepackage{enumitem}
\usepackage{color, colortbl}
\usepackage{xcolor}
\usepackage{tikz-dependency}
\usepackage{amsfonts}
\usepackage{comment}
\usepackage{tikz}

\usepackage{pifont}%

\definecolor{DarkGreen}{RGB}{1,180,32}
\newcommand{\cmark}{{\color{blue}\ding{51}}}
\newcommand{\xmark}{{\color{red}\ding{55}}}

\def\rot{\rotatebox}

\title{Probing LLMs for Multilingual Discourse Generalization \\ Through a Unified Label Set}

\author{\textbf{Florian Eichin}\thanks{Equal contribution.},  \textbf{Yang Janet Liu}\footnotemark[1], \textbf{Barbara Plank}, and \textbf{Michael A. Hedderich} \\
        MaiNLP, Center for Information and Language Processing, LMU Munich, Germany \\
        Munich Center for Machine Learning (MCML) \\
        \texttt{\{feichin,yliu,bplank,hedderich\}@cis.lmu.de}}

\begin{document}
\maketitle

\begin{abstract}

Discourse understanding is essential for many NLP tasks, yet most existing work remains constrained by framework-dependent discourse representations.
This work investigates whether large language models (LLMs) capture discourse knowledge that generalizes across languages and frameworks. We address this question along two dimensions: (1) developing a unified discourse relation label set to facilitate cross-lingual and cross-framework discourse analysis, and (2) probing LLMs to assess whether they encode generalizable discourse abstractions. 
Using multilingual discourse relation classification as a testbed, we examine a comprehensive set of $23$ LLMs of varying sizes and multilingual capabilities. Our results show that LLMs, especially those with multilingual training corpora, can generalize discourse information across languages and frameworks. Further layer-wise analyses reveal that language generalization at the discourse level is most salient in the intermediate layers. Lastly, our error analysis provides an account of challenging relation classes. 

\end{abstract}

\section{Introduction}
\label{sec:introduction}

Many approaches to NLP primarily focus on sentence-level analyses (e.g.~\citealt{heinzerling-strube-2019-sequence,pimentel-ryskina-etal-2021-sigmorphon,mrini-etal-2020-rethinking}).
However, there are many research questions which cannot be answered without considering sentences in a larger \textbf{discourse}: new meanings emerge from the relationships between sentences, and since more than one interpretation can be created, how do we determine the intended, most reasonable or justifiable meaning \cite{SchiffrinEtAl2015Ch0}? 

\begin{figure}[ht!]
    \centering
    \includegraphics*[trim=0 150 0 30,clip,width=\columnwidth]{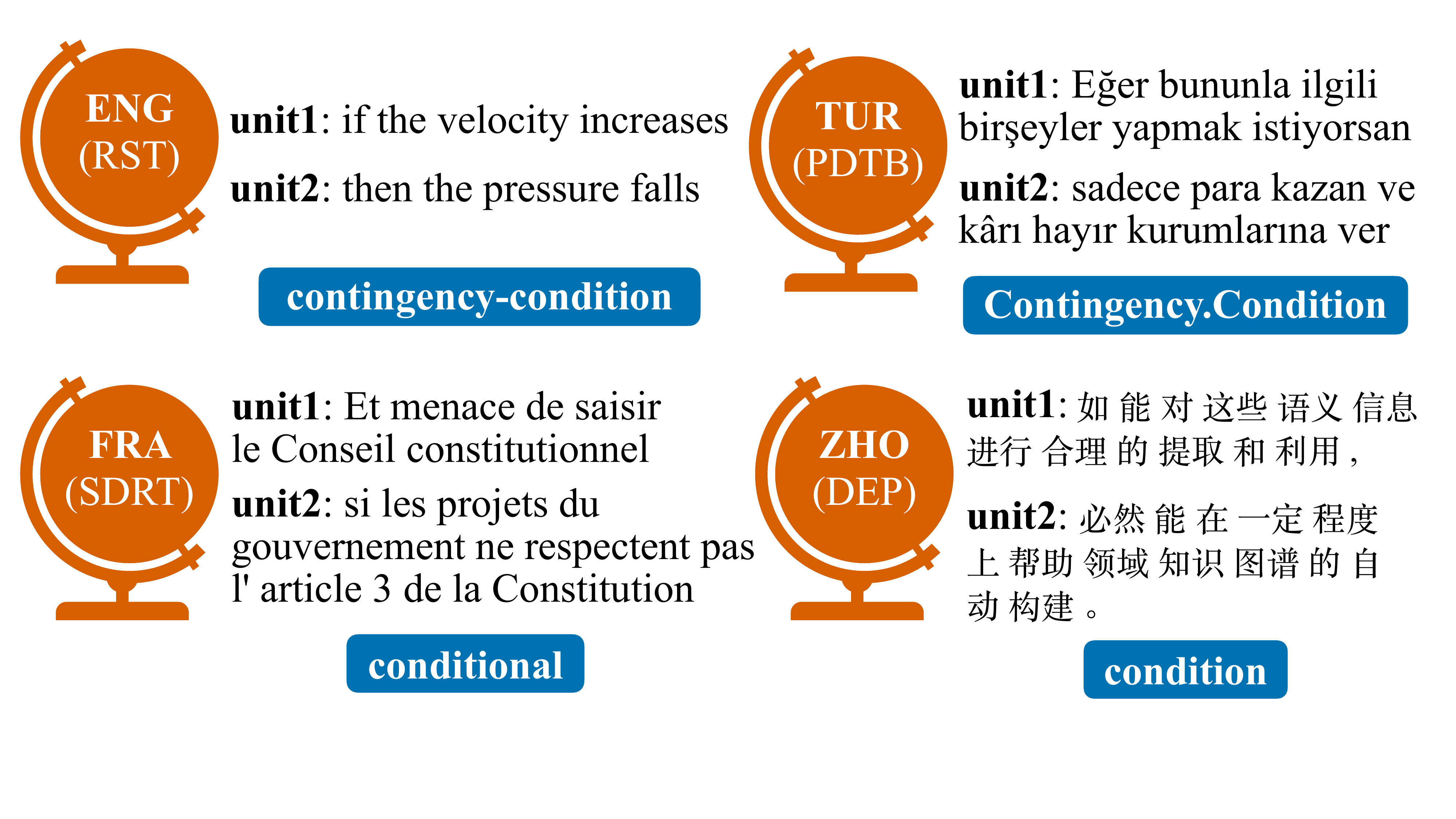}
    \vspace{-15pt}
    \caption{Examples of the core discourse relation \textsc{condition} \cite{Bunt2016ISOD} annotated in different frameworks and languages using different labels.}
    \vspace{-16pt}
    \label{fig:disco-rel-example}
\end{figure}

Despite significant progress in discourse processing \cite{WebberEtAl2024PDTB,ZeldesEtAl2024erst,stede2011discourse}, much of the research and resources remain constrained by theory-/framework-specific assumptions, limiting the generalizability of findings across languages, domains, and communicative intents \cite{liu-zeldes-2023-cant}. 
This leads to datasets that are tightly coupled to their respective frameworks and limits the development of generalizable discourse models. While there has been work on framework-dependent parsing that leverages resources from other frameworks \cite{braud-etal-2016-multi} or languages \cite{braud-etal-2017-cross,liu-etal-2021-dmrst}, the reliance on framework-specific corpora, which are typically scarce and skewed towards high-resource languages, further exacerbates the challenge of multilingual discourse processing. Thus, we need a unified view and approach to investigating discourse generalization.

From the \textbf{theory and data} perspectives, as argued in \citet{Bunt2016ISOD} and exemplified in \autoref{fig:disco-rel-example}, despite differences between frameworks, there exists a set of `core’ discourse relations which are commonly found in existing approaches to discourse relations and their annotation. 
From the \textbf{model} perspective, there is growing evidence demonstrating that large language models (LLMs) learn and share generalizable abstraction across typologically diverse languages (e.g.~\citealt{brinkmann2025largelanguagemodelsshare,peng-sogaard-2024-concept}), but such capabilities remain underexplored in discourse.

In this work, we address \textbf{discourse generalization} across two dimensions using discourse relation classification as a testbed: we first develop a unified discourse relation label set to enable cross-lingual and cross-framework discourse analysis on the the multilingual DISRPT benchmark \cite{braud-etal-2024-disrpt}. Then, we use \emph{probing} \citep{alain2018understandingintermediatelayersusing} to understand the internal discourse representations of $23$ LLMs with varying sizes and multilingual capabilities. We investigate whether their representations capture generalizable discourse abstractions across typologically diverse languages or whether they are limited by dataset-specific biases. 
We hypothesize that multilingual models might encode universal representations of certain relations while adjusting to language-specific features. 

We find that overall LLMs are able to generalize at the discourse level across languages and frameworks, and that multilingual training and larger model sizes both increase probe performance. Our layer-wise analyses show that language generalization at the discourse level is most salient in the intermediate layers, which are most predictive of multilingual discourse information. 
To our best knowledge, this is the first work to apply a unified label set to multilingual discourse relation classification at scale. 
Lastly, we discuss challenges and biases in LLM discourse representations, providing insights into the limitations and potential avenues for improving discourse modeling in multilingual and generalization settings. 
The implementations of our experiments are available on GitHub.\!\footnote{\url{https://github.com/mainlp/discourse_probes}}

\section{Related Work}
\label{sec:related-work}

\paragraph{Unifying Discourse Relations.} Discourse relations are fundamental to structuring coherent text and conveying meaning beyond the sentence level. Being able to identify and interpret these relations is crucial for many downstream NLP tasks, including machine comprehension \cite{narasimhan-barzilay-2015-machine,li-etal-2020-molweni}, sentiment analysis \cite{huber-carenini-2019-predicting}, question answering \cite{chai-jin-2004-discourse}, and summarization \cite{durrett-etal-2016-learning,cohan-etal-2018-discourse,xu-etal-2020-discourse,adams-etal-2023-generating}. However, due to different approaches to discourse relations, such as the Rhetorical Structure Theory (RST, \citealt{mann1988rhetorical}), the Penn Discourse Treenbank (PDTB, \citealt{webber2019penn}), the Segmented Discourse Representation Theory (SDRT, \citealt{asher2003logics}), Discourse Dependency Structure (DEP, \citealt{li-etal-2014-text,morey-etal-2018-dependency}), and the Cognitive Approach to Coherence Relations (CCR, \citealt{SandersSpoorenNoordman1992}), researchers have not reached a consensus on a unified set of discourse relations. 
There have been a few mapping proposals and examinations on existing annotations \cite{chiarcos-2014-towards,benamara-taboada-2015-mapping,Bunt2016ISOD,rehbein-etal-2016-annotating,SandersDembergHoekScholmanAsrZuffereyEversVermeul,Demberg2019HowCA}, but they are either merely focused on two frameworks at a time (e.g.~RST and SDRT in \citealt{benamara-taboada-2015-mapping}), or on high-resource languages and news-centric data such as mapping RST-DT and PDTB v2 by \citet{Demberg2019HowCA} and PDTB v3 by \citet{costa-etal-2023-mapping}. 

In particular, while \citet{Bunt2016ISOD} identified a set of core discourse relations, it did not cover DEP and was limited to English and French only. The examined corpora in their work also did not cover discourse phenomena concerning pragmatics or textual organization, both of which are indispensable aspects in discourse analysis. For instance, \textsc{background} and \textsc{motivation} are two RST-style relations that are not present in the examined RST Discourse Treebank (RST-DT, \citealt{CarlsonEtAl2003}). 
Both relations are expressed by various perlocutionary acts to affect readers' or speakers’ attitude and beliefs. 
To address these limitations, we first conduct an extensive review of previous work on discourse relation mapping proposals and 
present a unified label set (\S\ref{sec:data-unified-label}) to enable empirical studies in a multilingual setting, which has not been systematically explored before.

The 2021 DISRPT Shared Task \cite{zeldes-etal-2021-disrpt} introduced the first iteration of the discourse relation classification task in a unified format. It leveraged shared foundational assumptions across frameworks. 
However, no unified discourse relation labels were proposed, meaning that each dataset has its own label set, even for the ones that come from the same framework. We thus leverage this resource and propose a unified label set that is the first to be empirically tested in discourse relation classification across $13$ languages, four frameworks, and $26$ datasets, which cover various, modern genres, domains, and modalities.

\paragraph{Probing for Linguistic Representation and Generalization.}A growing body of research has explored the extent to which pretrained language models (PLMs) and LLMs encode linguistic representations and exhibit generalizable abstraction \cite{Hupkes2023}. They primarily focus on probing morphology \cite{brinkmann2025largelanguagemodelsshare}, syntax \cite{conneau-etal-2018-cram,hale-stanojevic-2024-llms}, semantic knowledge \cite{jumelet-etal-2021-language}, and syntax-semantics comprehension through cognitive linguistics paradigms such as construction grammar \cite{weissweiler-etal-2022-better}. 
While some studies demonstrate that LLMs share latent grammatical representations across diverse languages \cite{brinkmann2025largelanguagemodelsshare}, others also highlight key limitations in the semantic capabilities of LLMs \cite{scivetti2025assessinglanguagecomprehensionlarge}.

Previous work has examined the ability of PLMs/LLMs to understand discourse \cite{gan-etal-2024-assessing,saputa-etal-2024-polish,miao-etal-2024-discursive}, but their investigations are either limited to framework-dependent representations, monolingual datasets, or focus on single domains.
Specifically, \citet{koto-etal-2021-discourse} examined a variety of PLMs through a set of framework-dependent probing tasks for discourse coherence by looking at the residual stream (i.e.~token representations), while we approach discourse relation classification with a unified format and label set across various frameworks and $13$ languages (\S\ref{sec:data-unified-label}) using attentions (\S\ref{sec:methodology}), offering opportunities for investigating discourse generalization across languages and frameworks. 
\citet{kurfali-ostling-2021-probing} extended discourse probing to multilingual PLMs such as the multilingual BERT and XLM-RoBERTa \citep{devlin-etal-2019-bert,conneauUnsupervisedCrosslingualRepresentation2020} to examine how well they transfer discourse-level knowledge across languages, but their evaluation of discourse coherence was also framework-dependent and was only performed on two English datasets in the news domain. 
Lastly, \citet{kim-schuster-2023-entity} studied discourse understanding in LMs by probing their ability to track discourse entities, but their investigation is also limited to English.

\section{A Unified Label Set}
\label{sec:data-unified-label}

Building up on previous effort on mapping discourse relations across corpora and frameworks \cite{benamara-taboada-2015-mapping,Bunt2016ISOD,liu-zeldes-2023-cant}, we present a unified set of $17$ discourse relation classes to facilitate empirical investigation that is not constrained by framework-dependent discourse representations. 
This unified label set is motivated by both theoretical groundings and empirical studies, and takes annotation guidelines into considerations. Specifically, the proposed unified label set adapts the top-level classes from the mapping proposal described in \citet{benamara-taboada-2015-mapping} and extends it to phenomena frequent in dialogues such as acknowledgment, interruption, and correction \cite{asher-etal-2016-discourse}. 
Through a series of experiments, we demonstrate how our unified label set generalizes across languages and frameworks, providing a foundation for future empirical studies. 
Below we describe the top-level classes and include definitions and examples for all $17$ relation labels in Appendix \ref{app:labels}. 

\textbf{\textsc{temporal}} is mapped to framework-specific labels that establish a chronological sequence between events or states. Temporal relations indicate when one event occurs in relation to another such as \textit{before}, \textit{after}, or \textit{simultaneously}. These relations help organize discourse by providing a timeline of events. RST's \textsc{sequence}, PDTB's \textsc{Temporal.Asynchronous}/\textsc{Synchronous}, and SDRT's \textsc{temploc} and \textsc{flashback} (following \citealt{muller-etal-2012-constrained}) all fall under this class. 

\textbf{\textsc{structuring}} corresponds to fine-grained discourse relations that organize the structure of a text or conversation without necessarily conveying content-based meaning, connect discourse units of distinct context and equal prominence, and help guide the reader or listener through the discourse. RST-style relations such as \textsc{list} and \textsc{textual-organization}, PDTB's \textsc{expansion.disjunction},  and \textsc{parallel} and \textsc{alternation} in SDRT are mapped to this class. 

\textbf{\textsc{thematic}} is a broad class which includes relations among the content of the propositions, according to \citet{benamara-taboada-2015-mapping}. We adapt this top-level class to contain six subclasses: \textsc{framing}, \textsc{attribution}, \textsc{mode}, \textsc{reformulation}, \textsc{comparison}, and \textsc{elaboration}. In particular, we introduce \textsc{reformulation}, which corresponds to relations by which one discourse unit re-expresses the meaning of another in a different form and/or from a different perspective to help reinforce understanding. RST's \textsc{summary} and \textsc{restatement} and PDTB's \textsc{expansion.equivalence} are mapped to \textsc{reformulation}.

\begin{figure*}
    \centering
    \includegraphics*[trim=20 110 0 320,clip,width=0.95\linewidth]{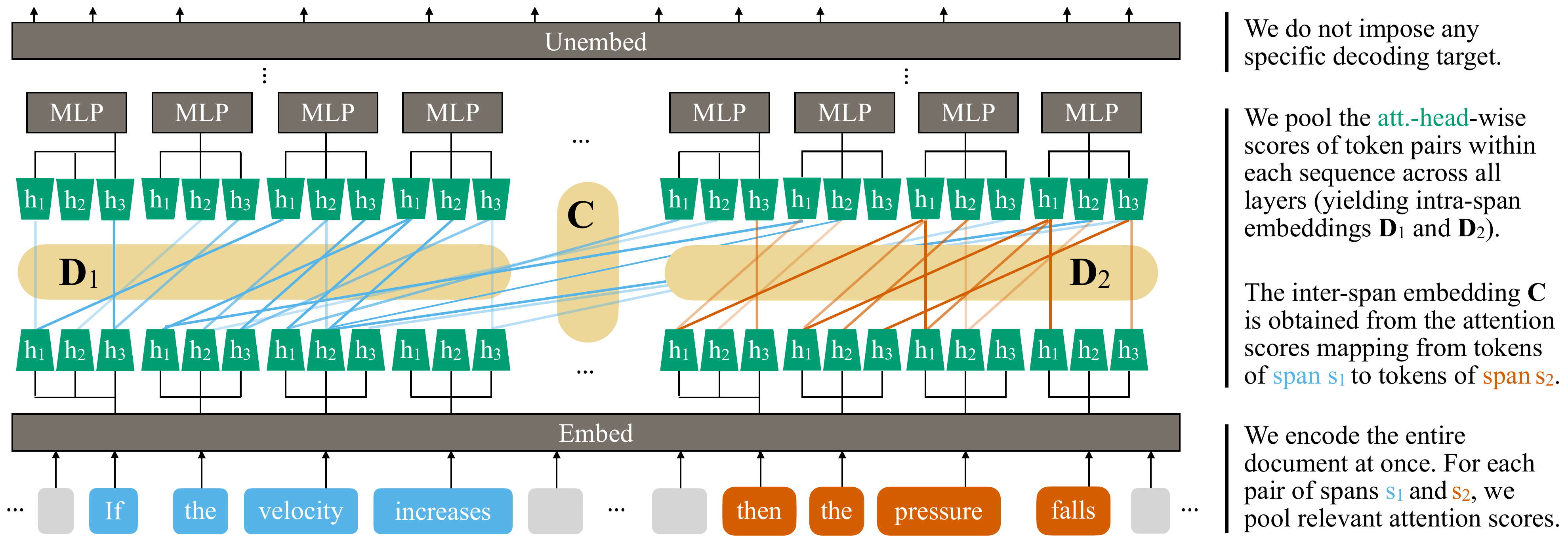}
    \caption{\textbf{Schematic visualization of how the attention representations are obtained.} We probe the concatenation of pooled representations $(\mathbf{D}_1, \mathbf{D}_2, \mathbf{C})$. Scores mapping from or to context-tokens are not considered. Note that we study decoder-only models where attention maps only to previous tokens.}
    \vspace{-10pt}
    \label{fig:viz-probing}
\end{figure*}

\textbf{\textsc{causal-argumentative}} contains subclasses that indicate a causal relation or involve rhetorical reasoning that shapes the coherence and persuasiveness of an argument: \textsc{causal}, \textsc{adversative}, \textsc{explanation}, \textsc{evaluation}, \textsc{contingency}, and \textsc{enablement}. \textsc{adversative} is defined as connecting discourse units for which some incompatibility is being highlighted and covers commonly used discourse relations such as \textsc{concession} and \textsc{contrast}.

\textbf{\textsc{topic-management}} contains three subclasses that cover discourse phenomena that involve interaction, topic shifts, and question-answer pairs: \textsc{topic-adjustment}, \textsc{topic-change}, and \textsc{topic-comment}. \textsc{topic-adjustment} is primarily used for cases where a discourse segment modifies, redirects, or adjusts the ongoing topic of discussion such as interruption, which signal deviations from the expected discourse progression, often reflecting interactive or dynamic aspects of communication.

\section{Probing LLMs for Discourse Relations}
\label{sec:methodology}

Discourse relation classification is the task of identifying the coherence relations that hold between different parts of a text, such as recognizing that one sentence specifies the cause of events in another, or that a subordinate clause indicates the condition of a main clause \cite{jm3nlp}, as exemplified in \autoref{fig:disco-rel-example}. 
Successfully solving this task requires combining the information contained in different parts of these sequences. We aim to investigate whether current LLMs have access to and process this information. Since almost all current state-of-the-art open-source models are decoder-only Transformer models \citep{2017nipsAttentionIsAllYouNeed}, we focus on this particular architecture.

In the computational graph of the Transformer, attention layers provide the only connections between the next token prediction and previous token positions of the input sequence. We therefore argue that generating predictions relying on discourse-level information necessarily has to involve the connections provided by the \textbf{self-attention} layers. To uncover the extent to which discourse information is represented, we thus propose to probe the attention scores between the tokens contained in the two input sequences for discourse relation classification.
Note that we do not assume that the processing of discourse-level phenomena is exclusively located in the attention layers. Rather, observing attention scores
provides a lightweight and scalable approach to our research question.

Probing requires a fixed length representation to enable training a classifier. %
Inspired by \citet{alain2018understandingintermediatelayersusing} and \citet{koto-etal-2021-discourse}, we use maximum pooling to turn the attention scores of variable token-length sentences into a fixed-length relational representation. Our approach is illustrated in \autoref{fig:viz-probing}.
To be more precise, for a document $d$ of length $N$ tokens, we propose to compute the full attention matrix $\mathbf{X} \in \mathbb{R}^{L \times H \times N \times N}$ of attention scores where $L$ is the number of layers and $H$ is the number of attention heads in each layer.
We do this by inputting the whole document $d$ into the model at once computing a single forward pass. Since we are only interested in the model internals, we ignore the outputs and thus do not impose any specific decoding strategy.
Let $I_1 = (i_1, i_2, ..., i_{N_1})$, $I_2 = (j_1, j_2, ..., j_{N_2})$ be the token indices of two sequences $s_1, s_2 \subset d$ where $I_1 < I_2$. The pooling step is carried out for each attention head in each layer. We thus have:
\begin{equation}
\begin{split}
	\mathbf{C} & = \left(\max(\{\mathbf{X}_{i, j, k, l} | k \in I_1, l\in I_2\})\right)_{i, j} \\
	\mathbf{D}_{m} & = \left(\max(\{\mathbf{X}_{i, j, k, l} | k \in I_m, l\in I_m\})\right)_{i, j} \\
\end{split}
\end{equation}
In other words, $\mathbf{C}, \mathbf{D}_{1}, \mathbf{D}_{2}$ are the matrices of maximum-pooled attention scores between and within the two spans respectively (we only consider the lower half of the attention matrix, as for decoder models the upper half is masked). We also ablate other strategies such as mean pooling and using subsets of the attention scores described (see \autoref{tab:disrpt-acc-by-model-ablated} and \autoref{tab:disrpt-acc-by-model-ablated2} in \autoref{app:analysis}), and find that the proposed setup strikes the best balance between performance and size of the representations. We probe the flattened concatenation 
\begin{equation}
\mathbf{a} = flatten((\mathbf{D}_1, \mathbf{D}_2, \mathbf{C}))\in \mathbb{R}^{3LH}
\end{equation}
Here, we follow \citet{tenneyWhatYouLearn2019} and use a two-layer MLP with $\mathrm{tanh}$ and Sigmoid activations $\sigma$. Our probe is thus a classifier of the form
\begin{equation}
\mathbf{y} = \sigma(\mathbf{W}_2\mathrm{tanh}(\mathbf{W}_1\mathbf{a} + \mathbf{b}_1)+\mathbf{b}_2)
\end{equation}
where $\mathbf{W}_1 \in \mathbb{R}^{D\times 3LH}, \mathbf{W}_2 \in \mathbb{R}^{C \times D}, \mathbf{b}_1 \in \mathbb{R}^D, \mathbf{b}_2 \in \mathbb{R}^C$ are parameters and $\mathbf{y}$ is the one-hot encoded target vector. The hidden size $D$ is a hyperparameter\ (more training details in \autoref{app:models-hyperparams}).

Since the memory requirements of this approach scale quadratically with the number of tokens, we restrict the maximum document length to $N_{max}=4000$, which is chosen to optimize the utilization of our GPUs. For documents with length $N>N_{max}$, we employ a moving window approach and encode slices of length $N_{max}$ at every stride $S=\frac{N_{max}}{2}$. Relations that span over parts of the document that cannot be captured by any of these windows are discarded and encoded by the mean of the remaining relations. In practice, this affects a fraction of less than $0.2\%$ of the DISRPT instances and thus has a negligible effect on the results.

Note that our approach requires only one forward pass per window encoding multiple relations in parallel and furthermore incorporates the document context into the representation.

We probe both the combined attention scores over all layers as well as the layer-wise representations. We hypothesize that the former will lead to better overall probe accuracy as the understanding of more complex discourse relations typically requires hierarchical processing steps. We include the latter layer-wise analysis to study the structural dynamics of discourse processing in the model.

\section{Experimental Setup}
\label{sec:experimental-setup}

\paragraph{Data.} We use the DISRPT benchmark from the DISRPT 2023 shared task \cite{braud-etal-2023-disrpt},\!\footnote{\url{https://github.com/disrpt/sharedtask2023}} a multilingual, multi-domain, and cross-framework (RST, PDTB, SDRT, and DEP) dataset covering $13$ languages from five language families \cite{wals}, %
with $224,281$ discourse relation instances from $23$ corpora (details in \autoref{app:data-disrpt}). While all annotations from all frameworks have been represented in a unified format, i.e., a set of discourse unit pairs for which a discourse relation is known to apply \cite{zeldes-etal-2021-disrpt}, the set of discourse relation labels is corpus-specific, preventing multilingual discourse analysis in a directly comparable and generalizable manner. Thus, we map corpus-specific labels in DISRPT to our proposed unified label set. 
\vspace{-2pt}
\paragraph{Models.} We study a wide range of decoder-only LLMs that reflect the diversity of the current state-of-the-art. Generally, we select models with publicly available weights along two dimensions: (1) the size of the model, and (2) the multilinguality of the training corpus. A full list of models including their advertised supported languages and parameter counts is shown in \autoref{tab:llms-overview} in Appendix \ref{app:models-hyperparams}.

\begin{figure*}
    \centering
    \includegraphics*[trim=40 50 50 285,clip,width=0.82\textwidth]{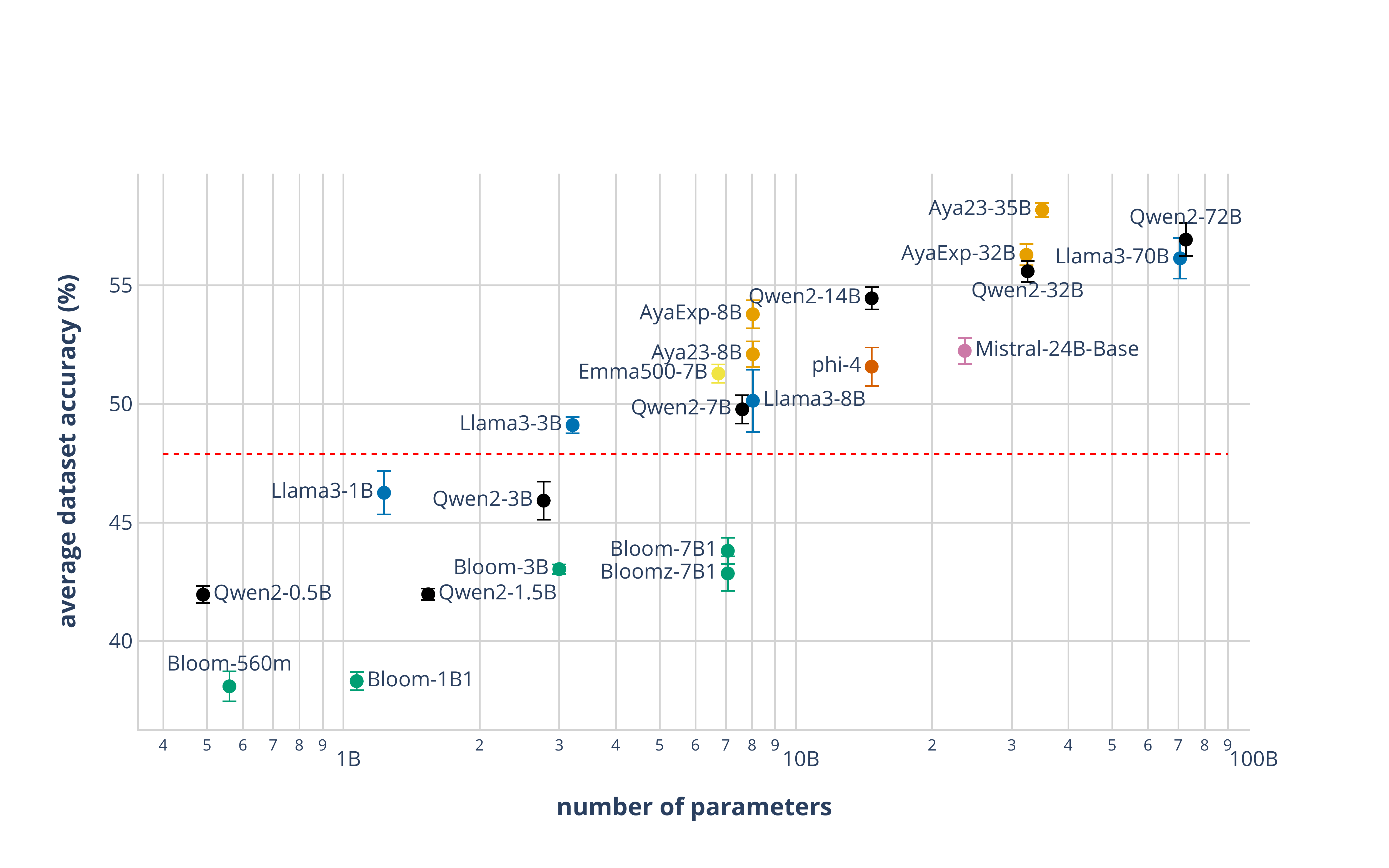}
    \caption{Mean accuracy over five runs of the probing classifiers trained on the entire DISRPT and full attention representations. The reference system DisCoDisCo achieved a mean accuracy of $47.9\%$ (the red dashed line).}
    \vspace{-8pt}
    \label{fig:param-acc-mlp-test}
\end{figure*}

Representing two of the most popular open-weights model families, we include most of the range of \texttt{Qwen2.5} \citep{qwenQwen25TechnicalReport2025} and \texttt{Llama3} \citep{grattafioriLlama3Herd2024} base models. The authors of both models report altered shares of multilingual data in their training corpora, with \texttt{Qwen2.5} `supporting' $77$\% and \texttt{Llama3} $54$\% of the languages included in DISRPT. Covering $62$\% and $100$\%, we include \texttt{Mistral-Small-24B} \citep{mistralMistralSmall32025} and \texttt{Emma500} \cite{jiEMMA500EnhancingMassively2024} as further recent multilingual models.
\texttt{BLOOM} is another model family that targets multilinguality \cite{scaoBLOOM176BParameterOpenAccess2023}, covering $54$\% of the languages in DISRPT. We include the smaller versions as well as the more recent \texttt{bloomz-7b1} model trained on additional data. Covering many languages in DISRPT with $85$\%, we include the \texttt{Aya-23}  \citep{aryabumiAya23Open2024} and \texttt{Aya-Expanse} \citep{dangAyaExpanseCombining2024} model families. Finally, we also include \texttt{Phi-4} \cite{abdinPhi4TechnicalReport2024} as a recent monolingual model. 

\vspace{-2pt}
\paragraph{DisCoDisCo.} To better contextualize the performance of our simple probes, we train DisCoDisCo \cite{gessler-etal-2021-discodisco}, the 2021 DISRPT shared task winning system for the discourse relation classification task \cite{zeldes-etal-2021-disrpt}. It was not tested during the 2023 edition, but the reported scores are better than the 2023 winning system on common corpora \cite{braud-etal-2023-disrpt}, justifying its use as a reference system. 
DisCoDisCo is a Transformer-based model consisting of a sentence-pair classifier. 
To ensure comparability, we train a single DisCoDisCo model on the entire DISRPT benchmark using \texttt{xlm-roberta-base} \cite{conneauUnsupervisedCrosslingualRepresentation2020} as a multilingual encoder. We also train dataset-specific models
following the setup in \citet{braud-etal-2023-disrpt}.
For all scenarios, we report both the dataset-wise accuracy and the mean accuracy.

\section{Results \& Analysis}
\label{sec:results-analysis}

\subsection{Overall Performance and Comparison}
\label{subsec:overall-perf-comp}

Figure \ref{fig:param-acc-mlp-test} shows the performance of our probes on DISRPT using the unified label set for the $23$ LLMs along with their sizes (see \autoref{tab:llms-overview} in Appendix \ref{app:models-hyperparams} for a detailed overview).
Overall, a clear trend emerges wherein larger models generally achieve higher accuracy, but with notable deviations based on their language coverage. Interestingly, even the \texttt{Llama3-3B} probe is able to outperform the fine-tuned DisCoDisCo trained on all languages ($49.1$\% vs $47.9$\%), while the larger model probes beat it by a margin of up to $10.3$\%.
Despite their multilingual training, the \texttt{BLOOM} model probes underperform compared to other LLMs and DisCoDisCo. Several factors may be attributed to this including smaller training data, tokenizer effect \cite{2022tokenizerImpact}, and limited coverage of languages present in DISRPT ($54$\%: French, Spanish, Portuguese, English, and Chinese). This is consistent and further supplements \citet{dakle2023understandingbloomempiricalstudy}'s findings on the evaluation of \texttt{BLOOM} models across a variety of syntactic and semantic tasks as well as their unsatisfying performance in multilingual settings. In addition, \texttt{Llama3} exhibits the lowest proportion of DISRPT languages covered, which does not seem to lead to a disadvantage as its probes outperform the similarly sized smaller versions of \texttt{Qwen2.5} and match its larger counterparts, reporting a higher coverage of `supported' languages.

Examining scaling trends, we observe a log-linear increase in probe performance across both the \texttt{Llama3} and \texttt{Qwen2.5} model families. Interestingly, the English-only \texttt{Phi-4} lags behind the similarly sized \texttt{Qwen2.5-14B} model, suggesting that multilingual training plays a critical role in multilingual discourse analysis. This further reinforces the importance of language coverage in training data, beyond simple model scaling, in achieving robust performance in discourse relation classification. Furthermore, despite being a fine-tuned version of \texttt{Llama2-7B}, \texttt{Emma500}’s multilingual training appears to give it an edge over the more recent and larger \texttt{Llama3-8B}.
Similarly, \texttt{Aya-23-35B}'s probe surpasses those of the largest \texttt{Qwen2.5-70B} and \texttt{Llama3-72B} models, despite operating with only half the number of parameters, emphasizing the efficiency of its learned representations. Given this evident edge in the overall probing performance, we focus the rest of our investigation into discourse generalization on \texttt{Aya-23-35B}.

\begin{figure*}[ht]
    \centering
    \includegraphics*[trim=0 0 50 305,clip,width=\textwidth]{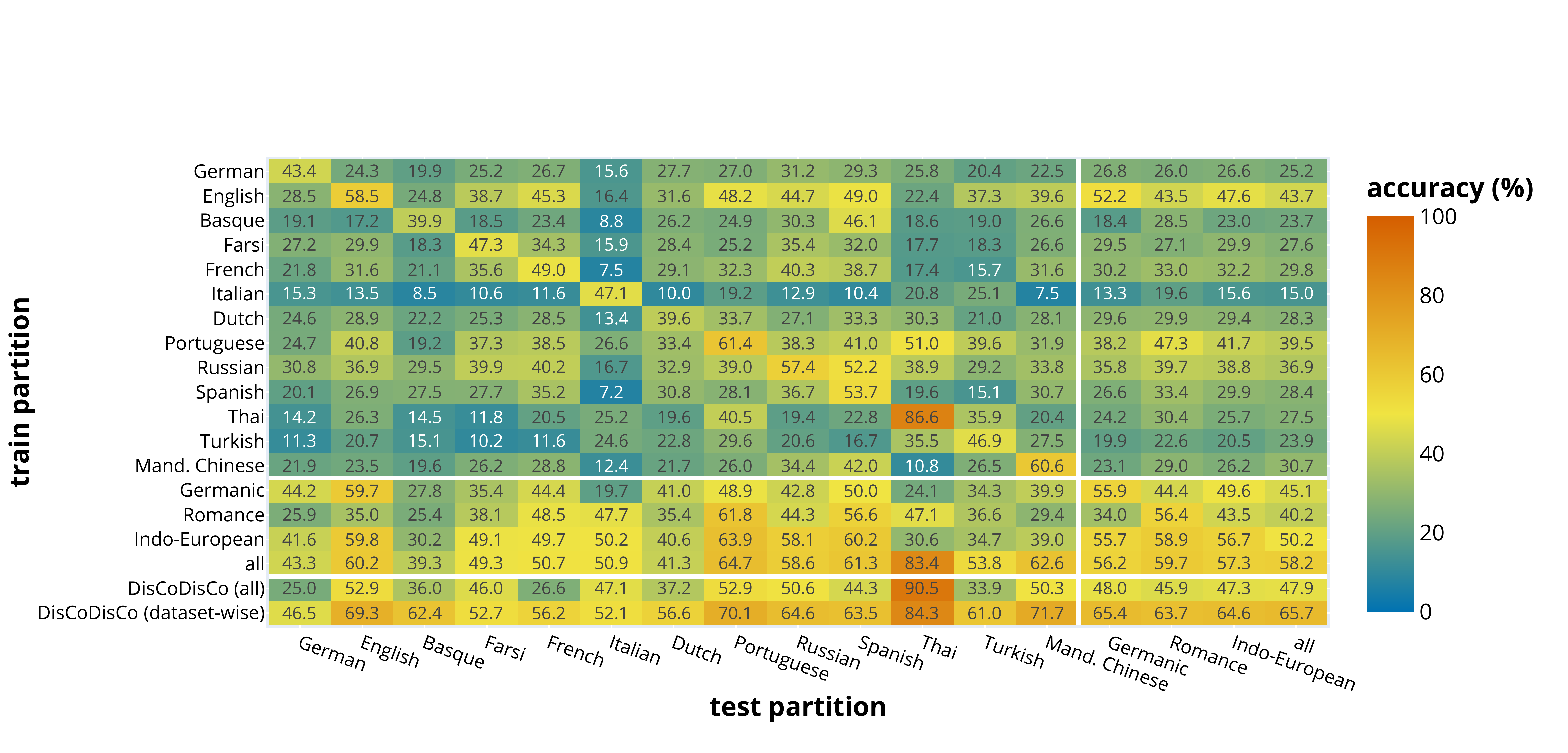}
    \vspace{-24pt}
    \caption{Mean accuracy over five runs of the \texttt{Aya-23-35B}-probe trained and tested on various partitions of DISRPT.}
    \vspace{-8pt}
    \label{fig:heatmap-language}
\end{figure*}

\subsection{Language Generalization}
\label{subsec:lang-gen}

We evaluate \texttt{Aya-23-35B}'s performance on the DISRPT test sets across $13$ languages and compare different training conditions: (1) monolingual training (\textbf{\textsc{mono-probe}}), (2) multilingual training with languages from the same language family (\textbf{\textsc{multi-lang-probe}}), and (3) multilingual training using instances from all languages (\textbf{\textsc{multi-all-probe}}). 
\autoref{fig:heatmap-language} shows the performance of the \texttt{Aya-23-35B} model trained and tested on all and subsets of the data given different training regimes. 

\paragraph{\textsc{multi-all-probe} \& \textsc{mono-probe}.} The probes trained on the entire DISRPT dataset match or outperform the monolingually trained probes across most test partitions, including the combined test set with all languages. This suggests that multilingual training benefits discourse relation classification, leveraging shared discourse patterns across languages. This is the opposite to what we see in the reference system DisCoDisCo, where the dataset-specific models outperform the model trained on all the data by a margin of $17.8$\%. 
For Farsi and Russian, the best performance is observed in the \textsc{multi-all-probe}, with $+2$\% and $+1.2$\% over their respective \textsc{mono-probe}s. The same generalization is observed for Turkish, Mandarin Chinese, and French, evidenced by an increase in accuracy of $6.9$\%, $2$\%, and $1.7$\% respectively.

\begin{figure*}[tbp]
    \centering
    \subfloat[languages and language groups]{\includegraphics*[trim=0 0 140 160,clip,width=0.32\textwidth]{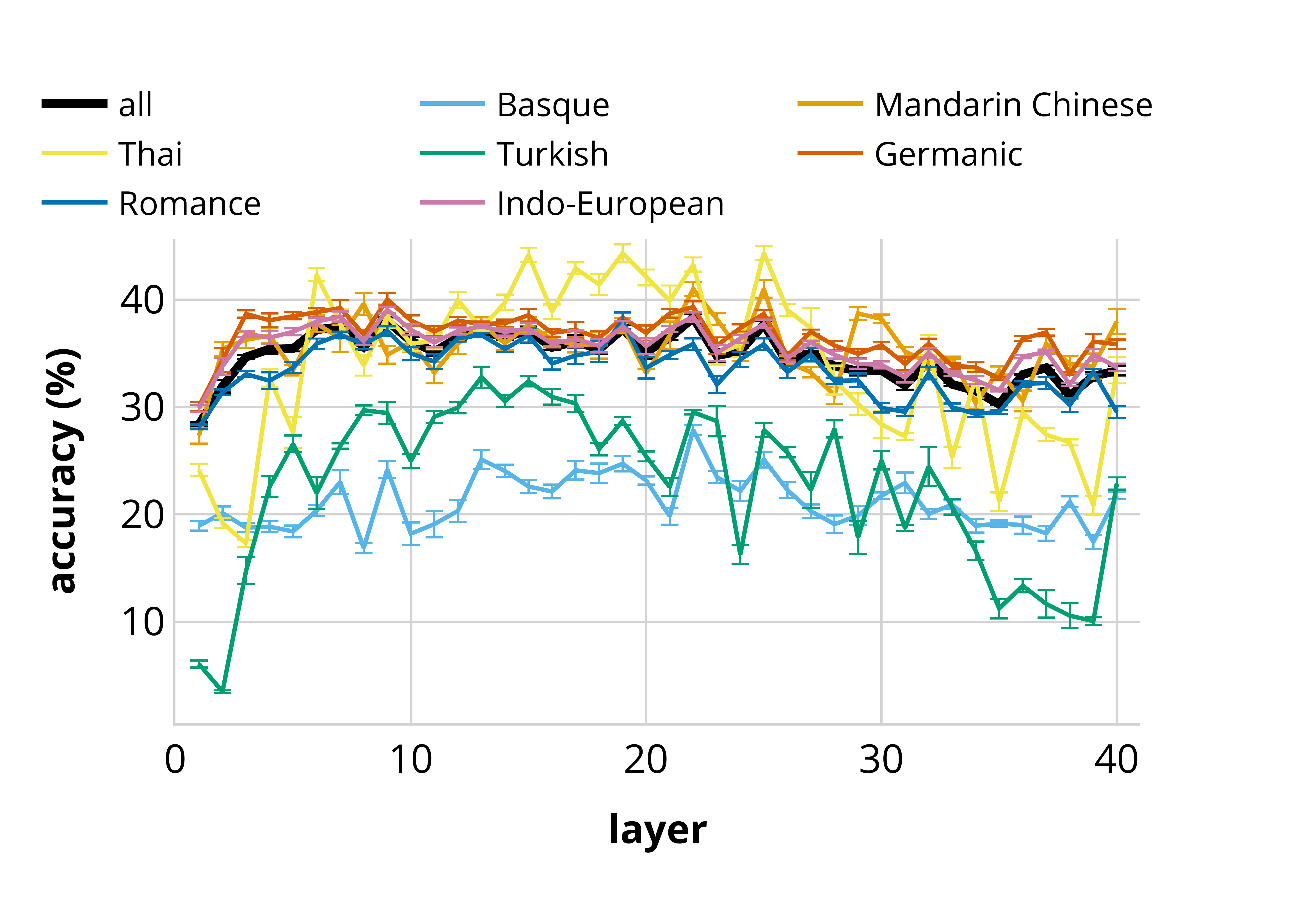}\label{fig:layer-wise-acc-lang}}
    \subfloat[top-level classes]{\includegraphics*[trim=0 0 140 160,clip,width=0.33\textwidth]{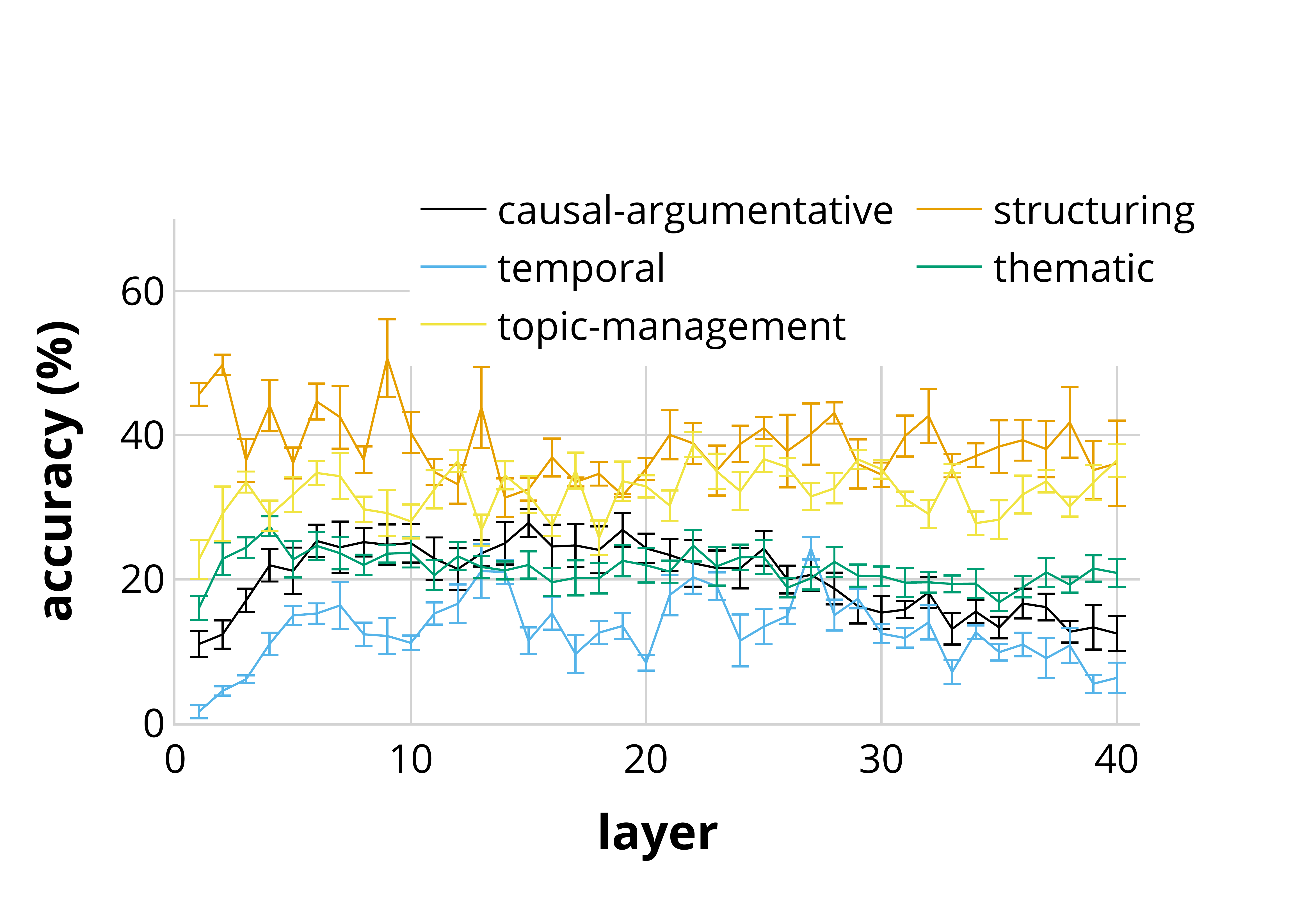}\label{fig:layer-wise-lab-acc-grouped}}
    \subfloat[\textsc{Causal-Argumentative}]{\includegraphics*[trim=0 0 140 160,clip,width=0.33\textwidth]{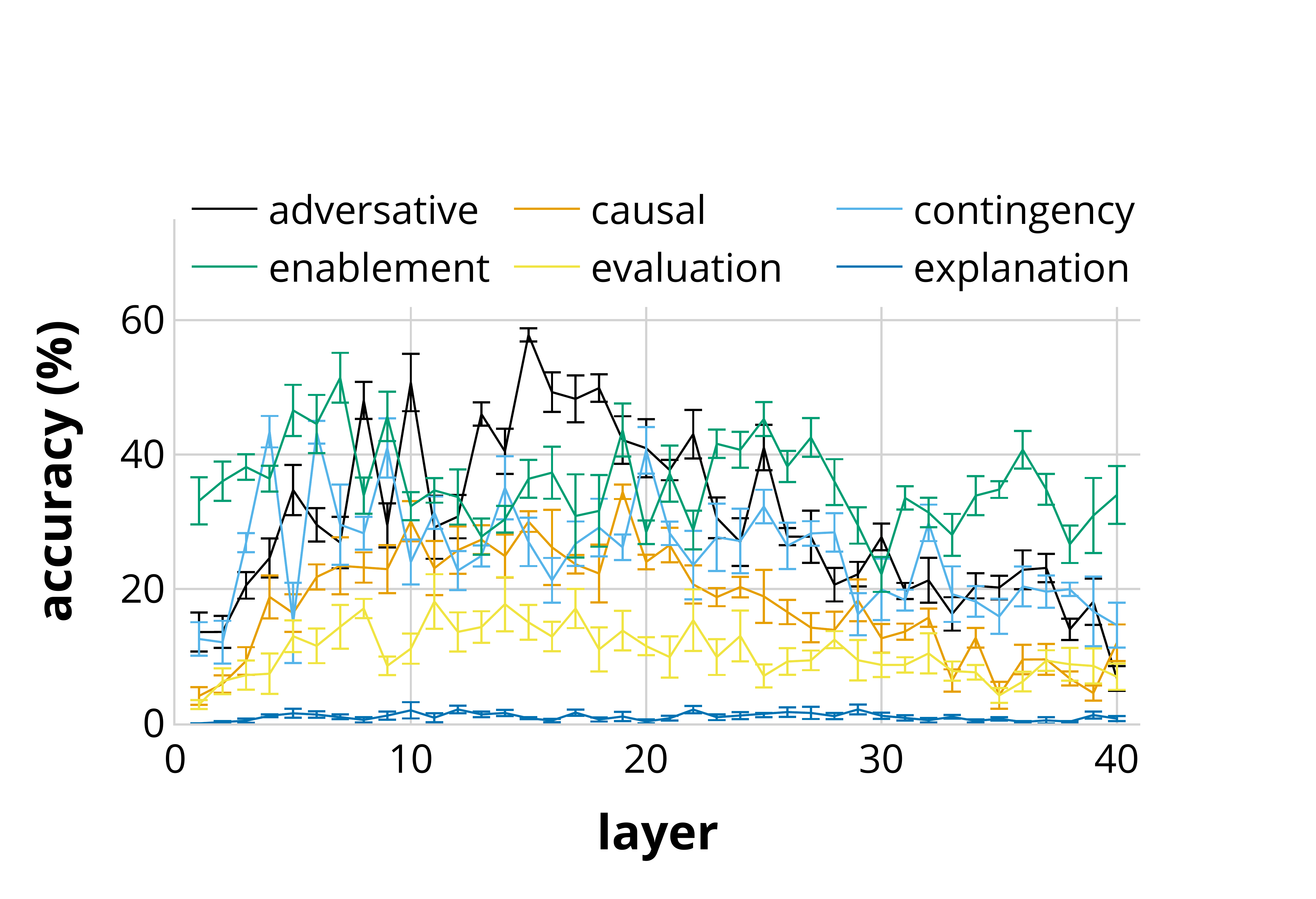}\label{fig:layer-wise-acc-causal-argu}}
    \caption{Layer-wise probe performance by languages and relation classes. Mean accuracy over five runs.} 
    \vspace{-4pt}
    \label{fig:layer-wise-acc}
\end{figure*}

There is one exception to this generalization trend: for the Thai dataset, the \textsc{mono-probe} outperforms \textsc{multi-all-probe} with a margin of $3.2\%$. 
This is likely due to the fact that the Thai dataset is relatively large and contains only news data and was only annotated with explicit discourse relations \cite{ThaiPDTB}, which has been reported to be considerably easier than implicit discourse relations \cite{knaebel-2021-discopy,braud-etal-2023-disrpt} due to the presence of connectives, which, while not unambiguous, help narrow down the likely senses of relations \cite{webber2019penn}. 
The probes for Basque consistently underperform, with the \textsc{multi-all-probe} achieving the same performance as the Basque-only probe. This is likely due to the fact that Basque is considered a language isolate, a language that has no demonstrable genetic relationship with any other language \cite{campbell2010language}. Besides, most models do not include it in their list of supported languages (see \autoref{tab:llms-overview} in \autoref{app:models-hyperparams}). The reference system DisCoDisCo also does not generalize well for Basque: its accuracy drops by $26.4$\% compared to the dataset-specific model.
Moreover, the \textsc{mono-probe}s for Basque and Turkish exhibit lower accuracy, which might mean that models trained on discourse tasks need more language-specific adaptations, particularly for morphologically rich and typologically different languages. 
\paragraph{\textsc{multi-lang-probe}.} Training on discourse relation instances from related languages often leads to better generalization than training on a single language. For instance, the \textsc{multi-lang-probe} for the Indo-European languages achieves reasonable performance across the Romance languages in DISRPT: there are improvements over the \textsc{mono-probe} for Portuguese ($+2.5$\%), Spanish ($+6.5$\%), and Italian ($+3.1$\%). 
For the Germanic and Romance language groups, we observe a similar generalization effect, though to a lesser extent. Notably, the Germanic \textsc{multi-lang-probe} leads to the highest accuracy on the German test set ($+0.8$\% over the \textsc{mono-probe}).
These are encouraging results as some of these languages do not have a large amount of instances in DISRPT: from the total number of data, Spanish covers $1.7$\%, German $1.19$\%, and Italian only $0.7$\% (see \autoref{tab:disrpt-overview} in \autoref{app:data-disrpt}). This suggests that leveraging discourse relation annotations from related language as well as the same underlying framework help with generalization. 
Adding to that, training on English only, which has the largest number of samples covering more than half of DISRPT, shows surprising generalization to other languages, with Portuguese and Spanish achieving an accuracy of $48.2\%$ and $49\%$ respectively, and French reaching $45.3\%$.

\subsection{Layer-wise Analysis}
\label{subsec:layer-wise-analysis}

The layer-wise probes reveal additional insights into the structural processing of discourse information within the model.
Firstly, Figure \ref{fig:layer-wise-acc-lang} shows the test set performance by language families. Here, we observe a trend by which the probes in lower to middle layers are the most performant as accuracy improves rapidly in the early layers and stabilizes around layer $10$–$15$. In particular, for Turkish and Thai, the probe accuracy drops by almost $50$\% in higher layers. This effect is also observed for the Chinese probe but to a smaller extent. This shows that the discourse representations are best aligned across languages in the middle layers, which could be explained by the findings of \citet{wendler-etal-2024-llamas} according to which multilingual computations are carried out in an English-aligned ``concept space'' in the intermediate layers.
This result is also aligned with concurrent work by \citet{skean2025layerlayeruncoveringhidden} which finds that final-layer representations are consistently outperformed by intermediate representations across a range of tasks and architectures. Our work supplements this by extending their finding to linguistic tasks such as discourse relation classification. 

Surprisingly, the final layer probes show another increase for Chinese, Turkish, and Thai, which is also reflected in the overall test set accuracy. This could indicate that their probes benefit from language-specific features, which might be more prevalent in layers close to the outputs if we assume a shared, compressed representation space in intermediate layers as proposed by \citet{wendler-etal-2024-llamas} and \citet{deletangLanguageModelingCompression2024}.

For the \textsc{multi-lang-probe}s of Romance, Germanic, and Indo-European, the corresponding performance is more constant, closely following the “all” curve (\textsc{multi-all-probe}). Here, the probes of the later layers show a drop of about $5$-$10$\% suggesting that their representations' alignment is more consistent throughout the model. This also indicates that both the higher and lower layers are involved in the discourse processing of LLMs.

Regarding relation classes, Figures \ref{fig:layer-wise-lab-acc-grouped} shows that \textsc{structuring} and \textsc{topic-management} achieve relatively higher accuracy across layers, peaking in the early-to-mid layers (around $5$-$15$), whereas \textsc{temporal} and \textsc{causal-argumentative} exhibit lower performance. Figure \ref{fig:layer-wise-acc-causal-argu} zooms in on the more fine-grained classes: \textsc{adversative} and \textsc{enablement} achieve the highest accuracy overall, with peaks around layers $10$-$20$, while \textsc{evaluation} and \textsc{explanation} remain consistently low. Layer-wise plots for other relation classes are provided in \autoref{fig:layer-wise-acc-rel-additional} in \autoref{app:analysis}. Overall, different relation classes appear to be encoded at varying layers, with some benefiting from middle-layer representations while others, particularly structure-oriented ones, maintain stable performance across layers. The drop in accuracy in later layers suggests that some discourse information might become less explicitly represented as the model progresses through deeper layers.

\subsection{Error Analysis}
\label{subsec:error-analysis}

\begin{figure}[ht!]
    \centering
    \vspace{6pt}
    \includegraphics*[trim=20 0 250 300,clip,width=\columnwidth]{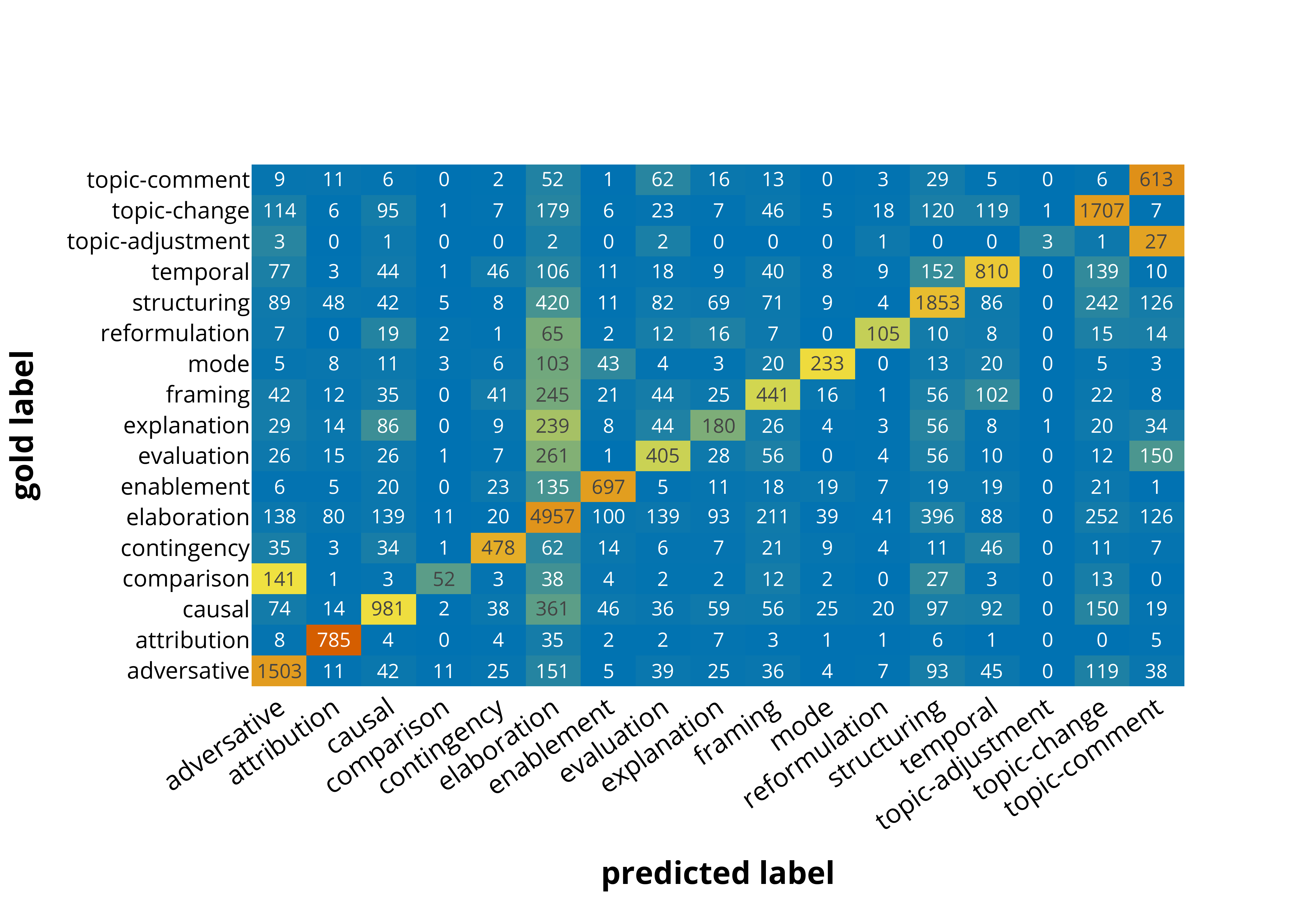}
    \vspace{-10pt}
    \caption{Confusion matrix of labels over \textsc{multi-all-probe} (colors normalized by row).}
    \vspace{-4pt}
    \label{fig:label-confmat}
\end{figure}

To better understand the probe performance, we plot a confusion matrix in \autoref{fig:label-confmat} to study systematic errors. Overall, \textsc{elaboration} is the most frequent label. Here, our probe has a bias of confusing it with labels such as \textsc{framing} and \textsc{explanation}. This suggests that the model struggles to differentiate between content expansion (\textsc{elaboration}) which sometimes overlaps semantically with \textsc{explanation}, and relations that involve additional information but with a primary focus on contextualization (\textsc{framing}). 
In other words, \textsc{elaboration} is mainly used to give additional information about an entity or a proposition, while \textsc{framing} provides information with a goal to increase a reader's understanding of an entity or a proposition \cite{mann1988rhetorical,carlson-marcu-01}. 
This corresponds to the distinction made by \citet{hovy1997parsimonious}: \textsc{elaboration} is considered an \textit{ideational} class that is used to express states of affairs in the world, not including the interlocutors; on the other hand, \textsc{framing} and \textsc{explanation} are considered \textit{interpersonal} and are expressed by various perlocutionary acts to affect readers' attitude and beliefs. 
\textsc{elaboration} is also a highly prevalent relation in general, making it a probable default prediction in ambiguous cases.

In addition, a majority of the \textsc{comparison} instances are predicted as \textsc{adversative}. A qualitative inspection reveals that this is due to some datasets not distinguishing similarities from differences, while the proposed unified label set does. We argue that similarity-based instances highlight commonalities between entities or situations and establish a shared property or behavior, reinforcing coherence by aligning elements. On the other hand, adversative-based cases emphasize differences, opposition, or unexpected alignments, which serve different pragmatic functions. Moreover, similarity-based comparisons often reinforce or extend prior discourse, leading to additive coherence; while adversative-based cases introduce shifts, which require the reader to re-evaluate assumptions and adjust interpretation. By distinguishing these two types, we think models can better capture rhetorical intent and argumentative structure.

\section{Conclusion}
\label{sec:conclusion}

Our study provides a comprehensive analysis of discourse generalization in LLMs, revealing important patterns in how these models encode discourse structures for cross-lingual transfer. 
To this end, we first present a unified discourse relation label set, which serves as a foundation for our probing experiments, allowing us to analyze discourse representations beyond individual frameworks. This is also the first work to apply a unified label set across frameworks and languages to multilingual discourse relation classification at scale. 

Our probes exhibit generalization, whereby training across languages generally outperforms language--specific probes, which is not the case for the reference system DisCoDisCo. Through multi-faceted analyses, we find that model size alone does not lead to discourse probing success; instead, multilingual training, dataset composition, and language-specific factors play significant roles. While larger models generally perform better, discrepancies such as the under-performing \texttt{BLOOM} and the best-performing \texttt{Aya-23-35B}-probe emphasize the importance of training data quality and architectural optimizations. Surprisingly, we find that some of our probes generalize to languages unseen during probe training. For example, the English \textsc{mono-probe} generalizes to Romance languages.
Furthermore, our layer-wise analysis suggests that discourse representations are best aligned across languages in the intermediate layers, with later layers refining these representations for specific relation types. In addition, models struggle with relations requiring implicit reasoning such as \textsc{explanation}.

Overall, our findings highlight the interplay between multilinguality, scaling, and internal representations of LLMs for multilingual discourse processing and provide insights into cross-lingual alignment of discourse relations. 
Understanding which discourse relations are well-captured by LLMs and which are not could help improve discourse parsing models by highlighting gaps in current representation. In addition, our findings may be useful for designing better discourse-aware pre-training or fine-tuning strategies. 
Future work can improve cross-lingual alignment by refining representations of challenging relation classes. 
More generally, our method provides a systematic way to investigate discourse representation encoded in LLMs, making it a useful tool for answering linguistic questions that can be formulated into a consecutive span representation. 
By advancing our understanding of discourse generalization in LLMs, we contribute towards more interpretable and robust NLP systems capable of nuanced language comprehension.

\section*{Limitations}
\label{sec:limitations}

While this work offers a first step towards understanding discourse generalization in LLMs across languages and discourse annotation frameworks, we acknowledge the following limitations related to data, unified representation, and methodological approach. 

Firstly, despite being multilingual, the DISRPT benchmark is imbalanced with regard to language coverage. English remains dominant (making up $53.5\%$ of the DISRPT benchmark, as shown in \autoref{tab:disrpt-overview}), which may impact the generalizability of our findings to lower-resource languages. To maintain comparability with prior work, we did not stratify the training data based on language sample sizes. However, we mitigated this by training language-specific probes (\textsc{mono-probe}) and \textsc{multi-lang-probe}s based on language families and reporting their respective performance in \autoref{fig:heatmap-language}. 
Similarly, while DISRPT includes multiple domains and genres, some are underrepresented such as conversational data or dialogues. This results in a label imbalance such as \textsc{topic-adjustment}, limiting statistical robustness in such cases. However, this also highlights the need for the creation of more balanced multilingual discourse resources. 

Secondly, our approach assumes a unified label set for discourse relations across languages and frameworks, which we present in \S\ref{sec:data-unified-label}. While this enables cross-linguistic and cross-framework discourse analysis, compromises were necessary, which to some extent simplify the complexity and granularity of discourse relations assumed by frameworks such as PDTB and RST. Fully standardizing and harmonizing discourse relations is inherently challenging, and finding a good trade-off between maintaining theoretical assumptions and ensuring practical applicability is crucial yet complex, as evidenced in previous mapping proposals and efforts \cite{benamara-taboada-2015-mapping,Demberg2019HowCA}. In addition, segmentation differences exist across frameworks, which can have an impact on the performance of our probes for certain relations such as \textsc{attribution}. 
This work should thus be interpreted with an awareness of the theoretical and practical difficulties in creating an informed and unified taxonomy suitable for both theoretical studies and computational research. This work should also be viewed as facilitating the development of a unified discourse relation representation for computational discourse modeling such as the effort and initiatives that have been put forward for dependency parsing (Universal Dependencies, \citealt{de-marneffe-etal-2021-universal}), empirical study of anaphora (Universal Anaphora, \citealt{poesio-etal-2024-universal}), and semantic parsing (Uniform Meaning Representation, \citealt{bonn-etal-2024-building}). 

Lastly, rather than using highly optimized architectures, we employed relatively simple probing methods, which aligns with our interest in assessing the intrinsic capabilities of LLMs for discourse processing. While achieving state-of-the-art performance was not our primary goal, better performance could likely be achieved by fine-tuning the LLMs. Future work would need to assess the trade-off between generalization and task-specific optimization.

\section*{Acknowledgments}
\label{sec:acknowledgments}

We would like to thank the members of the MaiNLP lab for their valuable feedback, especially to Philipp Mondorf, Siyao Peng, and Shijia Zhou.
We would also like to express our gratitude to Amir Zeldes for the constructive and valuable feedback. 
We thank the reviewers and the meta-reviewer for their feedback and suggestions. 
Lastly, we recognize the support for Yang Janet Liu and Barbara Plank through the ERC Consolidator Grant DIALECT 101043235.

\bibliography{custom,discoprob}  %

\begin{thebibliography}{82}
\providecommand{\natexlab}[1]{#1}

\bibitem[{Abdin et~al.(2024)Abdin, Aneja, Behl, Bubeck, Eldan, Gunasekar, Harrison, Hewett, Javaheripi, Kauffmann, Lee, Lee, Li, Liu, Mendes, Nguyen, Price, de~Rosa, Saarikivi, Salim, Shah, Wang, Ward, Wu, Yu, Zhang, and Zhang}]{abdinPhi4TechnicalReport2024}
Marah Abdin, Jyoti Aneja, Harkirat Behl, S{\'e}bastien Bubeck, Ronen Eldan, Suriya Gunasekar, Michael Harrison, Russell~J. Hewett, Mojan Javaheripi, Piero Kauffmann, James~R. Lee, Yin~Tat Lee, Yuanzhi Li, Weishung Liu, Caio C.~T. Mendes, Anh Nguyen, Eric Price, Gustavo de~Rosa, Olli Saarikivi, and 8 others. 2024.
\newblock \href {https://doi.org/10.48550/arXiv.2412.08905} {Phi-4 {{Technical Report}}}.
\newblock \emph{Preprint}, arXiv:2412.08905.

\bibitem[{Adams et~al.(2023)Adams, Fabbri, Ladhak, Elhadad, and McKeown}]{adams-etal-2023-generating}
Griffin Adams, Alex Fabbri, Faisal Ladhak, No{\'e}mie Elhadad, and Kathleen McKeown. 2023.
\newblock \href {https://doi.org/10.18653/v1/2023.acl-long.151} {Generating {EDU} extracts for plan-guided summary re-ranking}.
\newblock In \emph{Proceedings of the 61st Annual Meeting of the Association for Computational Linguistics (Volume 1: Long Papers)}, pages 2680--2697, Toronto, Canada. Association for Computational Linguistics.

\bibitem[{Alain and Bengio(2018)}]{alain2018understandingintermediatelayersusing}
Guillaume Alain and Yoshua Bengio. 2018.
\newblock \href {https://arxiv.org/abs/1610.01644} {Understanding intermediate layers using linear classifier probes}.
\newblock \emph{Preprint}, arXiv:1610.01644.

\bibitem[{Aryabumi et~al.(2024)Aryabumi, Dang, Talupuru, Dash, Cairuz, Lin, Venkitesh, Smith, Campos, Tan, Marchisio, Bartolo, Ruder, Locatelli, Kreutzer, Frosst, Gomez, Blunsom, Fadaee, {\"U}st{\"u}n, and Hooker}]{aryabumiAya23Open2024}
Viraat Aryabumi, John Dang, Dwarak Talupuru, Saurabh Dash, David Cairuz, Hangyu Lin, Bharat Venkitesh, Madeline Smith, Jon~Ander Campos, Yi~Chern Tan, Kelly Marchisio, Max Bartolo, Sebastian Ruder, Acyr Locatelli, Julia Kreutzer, Nick Frosst, Aidan Gomez, Phil Blunsom, Marzieh Fadaee, and 2 others. 2024.
\newblock \href {https://doi.org/10.48550/arXiv.2405.15032} {Aya 23: {{Open Weight Releases}} to {{Further Multilingual Progress}}}.
\newblock \emph{Preprint}, arXiv:2405.15032.

\bibitem[{Asher et~al.(2016)Asher, Hunter, Morey, Farah, and Afantenos}]{asher-etal-2016-discourse}
Nicholas Asher, Julie Hunter, Mathieu Morey, Benamara Farah, and Stergos Afantenos. 2016.
\newblock \href {https://aclanthology.org/L16-1432/} {Discourse structure and dialogue acts in multiparty dialogue: the {STAC} corpus}.
\newblock In \emph{Proceedings of the Tenth International Conference on Language Resources and Evaluation ({LREC}`16)}, pages 2721--2727, Portoro{\v{z}}, Slovenia. European Language Resources Association (ELRA).

\bibitem[{Asher and Lascarides(2003)}]{asher2003logics}
Nicholas Asher and Alex Lascarides. 2003.
\newblock \emph{Logics of {C}onversation}.
\newblock Cambridge University Press, Cambridge.

\bibitem[{Benamara and Taboada(2015)}]{benamara-taboada-2015-mapping}
Farah Benamara and Maite Taboada. 2015.
\newblock \href {https://doi.org/10.18653/v1/S15-1016} {Mapping different rhetorical relation annotations: A proposal}.
\newblock In \emph{Proceedings of the Fourth Joint Conference on Lexical and Computational Semantics}, pages 147--152, Denver, Colorado. Association for Computational Linguistics.

\bibitem[{Bonn et~al.(2024)Bonn, Buchholz, Chun, Cowell, Croft, Denk, Ge, Haji{\v{c}}, Lai, Martin, Myers, Palmer, Palmer, Post, Pustejovsky, Stenzel, Sun, Ure{\v{s}}ov{\'a}, Vallejos, Van~Gysel, Vigus, Xue, and Zhao}]{bonn-etal-2024-building}
Julia Bonn, Matthew~J. Buchholz, Jayeol Chun, Andrew Cowell, William Croft, Lukas Denk, Sijia Ge, Jan Haji{\v{c}}, Kenneth Lai, James~H. Martin, Skatje Myers, Alexis Palmer, Martha Palmer, Claire~Benet Post, James Pustejovsky, Kristine Stenzel, Haibo Sun, Zde{\v{n}}ka Ure{\v{s}}ov{\'a}, Rosa Vallejos, and 4 others. 2024.
\newblock \href {https://aclanthology.org/2024.lrec-main.229/} {Building a broad infrastructure for uniform meaning representations}.
\newblock In \emph{Proceedings of the 2024 Joint International Conference on Computational Linguistics, Language Resources and Evaluation (LREC-COLING 2024)}, pages 2537--2547, Torino, Italia. ELRA and ICCL.

\bibitem[{Braud et~al.(2017)Braud, Coavoux, and S{\o}gaard}]{braud-etal-2017-cross}
Chlo{\'e} Braud, Maximin Coavoux, and Anders S{\o}gaard. 2017.
\newblock \href {https://aclanthology.org/E17-1028/} {Cross-lingual {RST} discourse parsing}.
\newblock In \emph{Proceedings of the 15th Conference of the {E}uropean Chapter of the Association for Computational Linguistics: Volume 1, Long Papers}, pages 292--304, Valencia, Spain. Association for Computational Linguistics.

\bibitem[{Braud et~al.(2023)Braud, Liu, Metheniti, Muller, Rivi{\`e}re, Rutherford, and Zeldes}]{braud-etal-2023-disrpt}
Chlo{\'e} Braud, Yang~Janet Liu, Eleni Metheniti, Philippe Muller, Laura Rivi{\`e}re, Attapol Rutherford, and Amir Zeldes. 2023.
\newblock \href {https://doi.org/10.18653/v1/2023.disrpt-1.1} {The {DISRPT} 2023 shared task on elementary discourse unit segmentation, connective detection, and relation classification}.
\newblock In \emph{Proceedings of the 3rd Shared Task on Discourse Relation Parsing and Treebanking (DISRPT 2023)}, pages 1--21, Toronto, Canada. The Association for Computational Linguistics.

\bibitem[{Braud et~al.(2016)Braud, Plank, and S{\o}gaard}]{braud-etal-2016-multi}
Chlo{\'e} Braud, Barbara Plank, and Anders S{\o}gaard. 2016.
\newblock \href {https://aclanthology.org/C16-1179/} {Multi-view and multi-task training of {RST} discourse parsers}.
\newblock In \emph{Proceedings of {COLING} 2016, the 26th International Conference on Computational Linguistics: Technical Papers}, pages 1903--1913, Osaka, Japan. The COLING 2016 Organizing Committee.

\bibitem[{Braud et~al.(2024)Braud, Zeldes, Rivi{\`e}re, Liu, Muller, Sileo, and Aoyama}]{braud-etal-2024-disrpt}
Chlo{\'e} Braud, Amir Zeldes, Laura Rivi{\`e}re, Yang~Janet Liu, Philippe Muller, Damien Sileo, and Tatsuya Aoyama. 2024.
\newblock \href {https://aclanthology.org/2024.lrec-main.447/} {{DISRPT}: A multilingual, multi-domain, cross-framework benchmark for discourse processing}.
\newblock In \emph{Proceedings of the 2024 Joint International Conference on Computational Linguistics, Language Resources and Evaluation (LREC-COLING 2024)}, pages 4990--5005, Torino, Italia. ELRA and ICCL.

\bibitem[{Brinkmann et~al.(2025)Brinkmann, Wendler, Bartelt, and Mueller}]{brinkmann2025largelanguagemodelsshare}
Jannik Brinkmann, Chris Wendler, Christian Bartelt, and Aaron Mueller. 2025.
\newblock \href {https://aclanthology.org/2025.naacl-long.312/} {Large language models share representations of latent grammatical concepts across typologically diverse languages}.
\newblock In \emph{Proceedings of the 2025 Conference of the Nations of the Americas Chapter of the Association for Computational Linguistics: Human Language Technologies (Volume 1: Long Papers)}, pages 6131--6150, Albuquerque, New Mexico. Association for Computational Linguistics.

\bibitem[{Bunt and Prasad(2016)}]{Bunt2016ISOD}
Harry Bunt and Rashmi Prasad. 2016.
\newblock \href {https://api.semanticscholar.org/CorpusID:64709264} {{ISO DR-Core (ISO 24617-8): Core Concepts for the Annotation of Discourse Relations}}.
\newblock In \emph{Annual Meeting of the Association for Computational Linguistics}.

\bibitem[{Campbell(2010)}]{campbell2010language}
Lyle Campbell. 2010.
\newblock \href {https://journals.linguisticsociety.org/proceedings/index.php/BLS/article/view/3900/3629} {Language isolates and their history, or, what's weird, anyway?}
\newblock In \emph{Annual Meeting of the Berkeley Linguistics Society}, pages 16--31.

\bibitem[{Carlson and Marcu(2001)}]{carlson-marcu-01}
Lynn Carlson and Daniel Marcu. 2001.
\newblock \href {http://www.isi.edu/~marcu/discourse/tagging-ref-manual.pdf} {Discourse tagging reference manual}.
\newblock Technical Report ISI-TR-545, University of Southern California Information Sciences Institute.

\bibitem[{Carlson et~al.(2003)Carlson, Marcu, and Okurowski}]{CarlsonEtAl2003}
Lynn Carlson, Daniel Marcu, and Mary~Ellen Okurowski. 2003.
\newblock Building a {D}iscourse-{T}agged {C}orpus in the {F}ramework of {R}hetorical {S}tructure {T}heory.
\newblock In \emph{Current and New Directions in Discourse and Dialogue}, Text, Speech and Language Technology 22, pages 85--112. Kluwer, Dordrecht.

\bibitem[{Chai and Jin(2004)}]{chai-jin-2004-discourse}
Joyce~Y. Chai and Rong Jin. 2004.
\newblock \href {https://aclanthology.org/W04-2504/} {Discourse structure for context question answering}.
\newblock In \emph{Proceedings of the Workshop on Pragmatics of Question Answering at {HLT}-{NAACL} 2004}, pages 23--30, Boston, Massachusetts, USA. Association for Computational Linguistics.

\bibitem[{Chiarcos(2014)}]{chiarcos-2014-towards}
Christian Chiarcos. 2014.
\newblock \href {https://aclanthology.org/L14-1685/} {Towards interoperable discourse annotation. discourse features in the ontologies of linguistic annotation}.
\newblock In \emph{Proceedings of the Ninth International Conference on Language Resources and Evaluation ({LREC}`14)}, pages 4569--4577, Reykjavik, Iceland. European Language Resources Association (ELRA).

\bibitem[{Cohan et~al.(2018)Cohan, Dernoncourt, Kim, Bui, Kim, Chang, and Goharian}]{cohan-etal-2018-discourse}
Arman Cohan, Franck Dernoncourt, Doo~Soon Kim, Trung Bui, Seokhwan Kim, Walter Chang, and Nazli Goharian. 2018.
\newblock \href {https://doi.org/10.18653/v1/N18-2097} {A discourse-aware attention model for abstractive summarization of long documents}.
\newblock In \emph{Proceedings of the 2018 Conference of the North {A}merican Chapter of the Association for Computational Linguistics: Human Language Technologies, Volume 2 (Short Papers)}, pages 615--621, New Orleans, Louisiana. Association for Computational Linguistics.

\bibitem[{Conneau et~al.(2020)Conneau, Khandelwal, Goyal, Chaudhary, Wenzek, Guzm{\'a}n, Grave, Ott, Zettlemoyer, and Stoyanov}]{conneauUnsupervisedCrosslingualRepresentation2020}
Alexis Conneau, Kartikay Khandelwal, Naman Goyal, Vishrav Chaudhary, Guillaume Wenzek, Francisco Guzm{\'a}n, Edouard Grave, Myle Ott, Luke Zettlemoyer, and Veselin Stoyanov. 2020.
\newblock \href {https://doi.org/10.18653/v1/2020.acl-main.747} {Unsupervised {{Cross-lingual Representation Learning}} at {{Scale}}}.
\newblock In \emph{Proceedings of the 58th {{Annual Meeting}} of the {{Association}} for {{Computational Linguistics}}}, pages 8440--8451, Online. Association for Computational Linguistics.

\bibitem[{Conneau et~al.(2018)Conneau, Kruszewski, Lample, Barrault, and Baroni}]{conneau-etal-2018-cram}
Alexis Conneau, German Kruszewski, Guillaume Lample, Lo{\"i}c Barrault, and Marco Baroni. 2018.
\newblock \href {https://doi.org/10.18653/v1/P18-1198} {What you can cram into a single {\$}{\&}!{\#}* vector: Probing sentence embeddings for linguistic properties}.
\newblock In \emph{Proceedings of the 56th Annual Meeting of the Association for Computational Linguistics (Volume 1: Long Papers)}, pages 2126--2136, Melbourne, Australia. Association for Computational Linguistics.

\bibitem[{Costa et~al.(2023)Costa, Sheikh, and Kosseim}]{costa-etal-2023-mapping}
Nelson~Filipe Costa, Nadia Sheikh, and Leila Kosseim. 2023.
\newblock \href {https://aclanthology.org/2023.ranlp-1.39/} {Mapping explicit and implicit discourse relations between the {RST}-{DT} and the {PDTB} 3.0}.
\newblock In \emph{Proceedings of the 14th International Conference on Recent Advances in Natural Language Processing}, pages 344--352, Varna, Bulgaria. INCOMA Ltd., Shoumen, Bulgaria.

\bibitem[{Dakle et~al.(2023)Dakle, Rallabandi, and Raghavan}]{dakle2023understandingbloomempiricalstudy}
Parag~Pravin Dakle, SaiKrishna Rallabandi, and Preethi Raghavan. 2023.
\newblock \href {https://arxiv.org/abs/2211.14865} {{Understanding BLOOM: An empirical study on diverse NLP tasks}}.
\newblock \emph{Preprint}, arXiv:2211.14865.

\bibitem[{Dang et~al.(2024)Dang, Singh, D'souza, Ahmadian, Salamanca, Smith, Peppin, Hong, Govindassamy, Zhao, Kublik, Amer, Aryabumi, Campos, Tan, Kocmi, Strub, Grinsztajn, {Flet-Berliac}, Locatelli, Lin, Talupuru, Venkitesh, Cairuz, Yang, Chung, Ko, Shi, Shukayev, Bae, Piktus, Castagn{\'e}, {Cruz-Salinas}, Kim, {Crawhall-Stein}, Morisot, Roy, Blunsom, Zhang, Gomez, Frosst, Fadaee, Ermis, {\"U}st{\"u}n, and Hooker}]{dangAyaExpanseCombining2024}
John Dang, Shivalika Singh, Daniel D'souza, Arash Ahmadian, Alejandro Salamanca, Madeline Smith, Aidan Peppin, Sungjin Hong, Manoj Govindassamy, Terrence Zhao, Sandra Kublik, Meor Amer, Viraat Aryabumi, Jon~Ander Campos, Yi-Chern Tan, Tom Kocmi, Florian Strub, Nathan Grinsztajn, Yannis {Flet-Berliac}, and 26 others. 2024.
\newblock \href {https://doi.org/10.48550/arXiv.2412.04261} {Aya {{Expanse}}: {{Combining Research Breakthroughs}} for a {{New Multilingual Frontier}}}.
\newblock \emph{Preprint}, arXiv:2412.04261.

\bibitem[{de~Marneffe et~al.(2021)de~Marneffe, Manning, Nivre, and Zeman}]{de-marneffe-etal-2021-universal}
Marie-Catherine de~Marneffe, Christopher~D. Manning, Joakim Nivre, and Daniel Zeman. 2021.
\newblock \href {https://doi.org/10.1162/coli_a_00402} {Universal dependencies}.
\newblock \emph{Computational Linguistics}, 47(2):255--308.

\bibitem[{Deletang et~al.(2024)Deletang, Ruoss, Duquenne, Catt, Genewein, Mattern, Grau-Moya, Wenliang, Aitchison, Orseau, Hutter, and Veness}]{deletangLanguageModelingCompression2024}
Gregoire Deletang, Anian Ruoss, Paul-Ambroise Duquenne, Elliot Catt, Tim Genewein, Christopher Mattern, Jordi Grau-Moya, Li~Kevin Wenliang, Matthew Aitchison, Laurent Orseau, Marcus Hutter, and Joel Veness. 2024.
\newblock \href {https://openreview.net/forum?id=jznbgiynus} {Language modeling is compression}.
\newblock In \emph{The Twelfth International Conference on Learning Representations}.

\bibitem[{Demberg et~al.(2019)Demberg, Scholman, and Asr}]{Demberg2019HowCA}
Vera Demberg, Merel Scholman, and Fatemeh~Torabi Asr. 2019.
\newblock \href {https://journals.uic.edu/ojs/index.php/dad/article/view/10694} {{How compatible are our discourse annotation frameworks? Insights from mapping RST-DT and PDTB annotations}}.
\newblock \emph{Dialogue \& Discourse}, 10:87--135.

\bibitem[{Devlin et~al.(2019)Devlin, Chang, Lee, and Toutanova}]{devlin-etal-2019-bert}
Jacob Devlin, Ming-Wei Chang, Kenton Lee, and Kristina Toutanova. 2019.
\newblock \href {https://doi.org/10.18653/v1/N19-1423} {{BERT}: Pre-training of deep bidirectional transformers for language understanding}.
\newblock In \emph{Proceedings of the 2019 Conference of the North {A}merican Chapter of the Association for Computational Linguistics: Human Language Technologies, Volume 1 (Long and Short Papers)}, pages 4171--4186, Minneapolis, Minnesota. Association for Computational Linguistics.

\bibitem[{Dryer and Haspelmath(2013)}]{wals}
Matthew~S. Dryer and Martin Haspelmath, editors. 2013.
\newblock \href {https://doi.org/10.5281/zenodo.13950591} {\emph{WALS Online (v2020.4)}}.
\newblock Zenodo.

\bibitem[{Durrett et~al.(2016)Durrett, Berg-Kirkpatrick, and Klein}]{durrett-etal-2016-learning}
Greg Durrett, Taylor Berg-Kirkpatrick, and Dan Klein. 2016.
\newblock \href {https://doi.org/10.18653/v1/P16-1188} {Learning-based single-document summarization with compression and anaphoricity constraints}.
\newblock In \emph{Proceedings of the 54th Annual Meeting of the Association for Computational Linguistics (Volume 1: Long Papers)}, pages 1998--2008, Berlin, Germany. Association for Computational Linguistics.

\bibitem[{Gan et~al.(2024)Gan, Poesio, and Yu}]{gan-etal-2024-assessing}
Yujian Gan, Massimo Poesio, and Juntao Yu. 2024.
\newblock \href {https://aclanthology.org/2024.lrec-main.145/} {Assessing the capabilities of large language models in coreference: An evaluation}.
\newblock In \emph{Proceedings of the 2024 Joint International Conference on Computational Linguistics, Language Resources and Evaluation (LREC-COLING 2024)}, pages 1645--1665, Torino, Italia. ELRA and ICCL.

\bibitem[{Gessler et~al.(2021)Gessler, Behzad, Liu, Peng, Zhu, and Zeldes}]{gessler-etal-2021-discodisco}
Luke Gessler, Shabnam Behzad, Yang~Janet Liu, Siyao Peng, Yilun Zhu, and Amir Zeldes. 2021.
\newblock \href {https://doi.org/10.18653/v1/2021.disrpt-1.6} {{D}is{C}o{D}is{C}o at the {DISRPT}2021 shared task: A system for discourse segmentation, classification, and connective detection}.
\newblock In \emph{Proceedings of the 2nd Shared Task on Discourse Relation Parsing and Treebanking (DISRPT 2021)}, pages 51--62, Punta Cana, Dominican Republic. Association for Computational Linguistics.

\bibitem[{Grattafiori et~al.(2024)Grattafiori, Dubey, Jauhri, Pandey, Kadian, {Al-Dahle}, Letman, Mathur, Schelten, Vaughan, Yang, Fan, Goyal, Hartshorn, Yang, Mitra, Sravankumar, Korenev, Hinsvark, Rao, Zhang, Rodriguez, Gregerson, Spataru, Roziere, Biron, Tang, Chern, Caucheteux, Nayak, Bi, Marra, McConnell, Keller, Touret, Wu, Wong, Ferrer, Nikolaidis, Allonsius, Song, Pintz, Livshits, Wyatt, Esiobu, Choudhary, Mahajan, {Garcia-Olano}, Perino, Hupkes, Lakomkin, AlBadawy, Lobanova, Dinan, Smith, Radenovic, Guzm{\'a}n, Zhang, Synnaeve, Lee, Anderson, Thattai, Nail, Mialon, Pang, Cucurell, Nguyen, Korevaar, Xu, Touvron, Zarov, Ibarra, Kloumann, Misra, Evtimov, Zhang, Copet, Lee, Geffert, Vranes, Park, Mahadeokar, Shah, van~der Linde, Billock, Hong, Lee, Fu, Chi, Huang, Liu, Wang, Yu, Bitton, Spisak, Park, Rocca, Johnstun, Saxe, Jia, Alwala, Prasad, Upasani, Plawiak, Li, Heafield, Stone, {El-Arini}, Iyer, Malik, Chiu, Bhalla, Lakhotia, {Rantala-Yeary}, van~der Maaten, Chen, Tan, Jenkins, Martin, Madaan, Malo,
  Blecher, Landzaat, de~Oliveira, Muzzi, Pasupuleti, Singh, Paluri, Kardas, Tsimpoukelli, Oldham, Rita, Pavlova, Kambadur, Lewis, Si, Singh, Hassan, Goyal, Torabi, Bashlykov, Bogoychev, Chatterji, Zhang, Duchenne, {\c C}elebi, Alrassy, Zhang, Li, Vasic, Weng, Bhargava, Dubal, Krishnan, Koura, Xu, He, Dong, Srinivasan, Ganapathy, Calderer, Cabral, Stojnic, Raileanu, Maheswari, Girdhar, Patel, Sauvestre, Polidoro, Sumbaly, Taylor, Silva, Hou, Wang, Hosseini, Chennabasappa, Singh, Bell, Kim, Edunov, Nie, Narang, Raparthy, Shen, Wan, Bhosale, Zhang, Vandenhende, Batra, Whitman, Sootla, Collot, Gururangan, Borodinsky, Herman, Fowler, Sheasha, Georgiou, Scialom, Speckbacher, Mihaylov, Xiao, Karn, Goswami, Gupta, Ramanathan, Kerkez, Gonguet, Do, Vogeti, Albiero, Petrovic, Chu, Xiong, Fu, Meers, Martinet, Wang, Wang, Tan, Xia, Xie, Jia, Wang, Goldschlag, Gaur, Babaei, Wen, Song, Zhang, Li, Mao, Coudert, Yan, Chen, Papakipos, Singh, Srivastava, Jain, Kelsey, Shajnfeld, Gangidi, Victoria, Goldstand, Menon, Sharma,
  Boesenberg, Baevski, Feinstein, Kallet, Sangani, Teo, Yunus, Lupu, Alvarado, Caples, Gu, Ho, Poulton, Ryan, Ramchandani, Dong, Franco, Goyal, Saraf, Chowdhury, Gabriel, Bharambe, Eisenman, Yazdan, James, Maurer, Leonhardi, Huang, Loyd, Paola, Paranjape, Liu, Wu, Ni, Hancock, Wasti, Spence, Stojkovic, Gamido, Montalvo, Parker, Burton, Mejia, Liu, Wang, Kim, Zhou, Hu, Chu, Cai, Tindal, Feichtenhofer, Gao, Civin, Beaty, Kreymer, Li, Adkins, Xu, Testuggine, David, Parikh, Liskovich, Foss, Wang, Le, Holland, Dowling, Jamil, Montgomery, Presani, Hahn, Wood, Le, Brinkman, Arcaute, Dunbar, Smothers, Sun, Kreuk, Tian, Kokkinos, Ozgenel, Caggioni, Kanayet, Seide, Florez, Schwarz, Badeer, Swee, Halpern, Herman, Sizov, Guangyi, Zhang, Lakshminarayanan, Inan, Shojanazeri, Zou, Wang, Zha, Habeeb, Rudolph, Suk, Aspegren, Goldman, Zhan, Damlaj, Molybog, Tufanov, Leontiadis, Veliche, Gat, Weissman, Geboski, Kohli, Lam, Asher, Gaya, Marcus, Tang, Chan, Zhen, Reizenstein, Teboul, Zhong, Jin, Yang, Cummings, Carvill, Shepard,
  McPhie, Torres, Ginsburg, Wang, Wu, U, Saxena, Khandelwal, Zand, Matosich, Veeraraghavan, Michelena, Li, Jagadeesh, Huang, Chawla, Huang, Chen, Garg, A, Silva, Bell, Zhang, Guo, Yu, Moshkovich, Wehrstedt, Khabsa, Avalani, Bhatt, Mankus, Hasson, Lennie, Reso, Groshev, Naumov, Lathi, Keneally, Liu, Seltzer, Valko, Restrepo, Patel, Vyatskov, Samvelyan, Clark, Macey, Wang, Hermoso, Metanat, Rastegari, Bansal, Santhanam, Parks, White, Bawa, Singhal, Egebo, Usunier, Mehta, Laptev, Dong, Cheng, Chernoguz, Hart, Salpekar, Kalinli, Kent, Parekh, Saab, Balaji, Rittner, Bontrager, Roux, Dollar, Zvyagina, Ratanchandani, Yuvraj, Liang, Alao, Rodriguez, Ayub, Murthy, Nayani, Mitra, Parthasarathy, Li, Hogan, Battey, Wang, Howes, Rinott, Mehta, Siby, Bondu, Datta, Chugh, Hunt, Dhillon, Sidorov, Pan, Mahajan, Verma, Yamamoto, Ramaswamy, Lindsay, Lindsay, Feng, Lin, Zha, Patil, Shankar, Zhang, Zhang, Wang, Agarwal, Sajuyigbe, Chintala, Max, Chen, Kehoe, Satterfield, Govindaprasad, Gupta, Deng, Cho, Virk, Subramanian,
  Choudhury, Goldman, Remez, Glaser, Best, Koehler, Robinson, Li, Zhang, Matthews, Chou, Shaked, Vontimitta, Ajayi, Montanez, Mohan, Kumar, Mangla, Ionescu, Poenaru, Mihailescu, Ivanov, Li, Wang, Jiang, Bouaziz, Constable, Tang, Wu, Wang, Wu, Gao, Kleinman, Chen, Hu, Jia, Qi, Li, Zhang, Zhang, Adi, Nam, Yu, Wang, Zhao, Hao, Qian, Li, He, Rait, DeVito, Rosnbrick, Wen, Yang, Zhao, and Ma}]{grattafioriLlama3Herd2024}
Aaron Grattafiori, Abhimanyu Dubey, Abhinav Jauhri, Abhinav Pandey, Abhishek Kadian, Ahmad {Al-Dahle}, Aiesha Letman, Akhil Mathur, Alan Schelten, Alex Vaughan, Amy Yang, Angela Fan, Anirudh Goyal, Anthony Hartshorn, Aobo Yang, Archi Mitra, Archie Sravankumar, Artem Korenev, Arthur Hinsvark, and 542 others. 2024.
\newblock \href {https://doi.org/10.48550/arXiv.2407.21783} {The {{Llama}} 3 {{Herd}} of {{Models}}}.
\newblock \emph{Preprint}, arXiv:2407.21783.

\bibitem[{Hale and Stanojevi{\'c}(2024)}]{hale-stanojevic-2024-llms}
John~T. Hale and Milo{\v{s}} Stanojevi{\'c}. 2024.
\newblock \href {https://doi.org/10.18653/v1/2024.emnlp-main.950} {Do {LLM}s learn a true syntactic universal?}
\newblock In \emph{Proceedings of the 2024 Conference on Empirical Methods in Natural Language Processing}, pages 17106--17119, Miami, Florida, USA. Association for Computational Linguistics.

\bibitem[{Heinzerling and Strube(2019)}]{heinzerling-strube-2019-sequence}
Benjamin Heinzerling and Michael Strube. 2019.
\newblock \href {https://doi.org/10.18653/v1/P19-1027} {Sequence tagging with contextual and non-contextual subword representations: A multilingual evaluation}.
\newblock In \emph{Proceedings of the 57th Annual Meeting of the Association for Computational Linguistics}, pages 273--291, Florence, Italy. Association for Computational Linguistics.

\bibitem[{Hovy and Maier(1997)}]{hovy1997parsimonious}
Eduard~H Hovy and Elisabeth Maier. 1997.
\newblock Parsimonious or {P}rofligate: {H}ow {M}any and {W}hich {D}iscourse {S}tructure {R}elations.
\newblock \emph{Discourse Processes}.

\bibitem[{Huber and Carenini(2019)}]{huber-carenini-2019-predicting}
Patrick Huber and Giuseppe Carenini. 2019.
\newblock \href {https://doi.org/10.18653/v1/D19-1235} {Predicting discourse structure using distant supervision from sentiment}.
\newblock In \emph{Proceedings of the 2019 Conference on Empirical Methods in Natural Language Processing and the 9th International Joint Conference on Natural Language Processing (EMNLP-IJCNLP)}, pages 2306--2316, Hong Kong, China. Association for Computational Linguistics.

\bibitem[{Hupkes et~al.(2023)Hupkes, Giulianelli, Dankers, Artetxe, Elazar, Pimentel, Christodoulopoulos, Lasri, Saphra, Sinclair, Ulmer, Schottmann, Batsuren, Sun, Sinha, Khalatbari, Ryskina, Frieske, Cotterell, and Jin}]{Hupkes2023}
Dieuwke Hupkes, Mario Giulianelli, Verna Dankers, Mikel Artetxe, Yanai Elazar, Tiago Pimentel, Christos Christodoulopoulos, Karim Lasri, Naomi Saphra, Arabella Sinclair, Dennis Ulmer, Florian Schottmann, Khuyagbaatar Batsuren, Kaiser Sun, Koustuv Sinha, Leila Khalatbari, Maria Ryskina, Rita Frieske, Ryan Cotterell, and Zhijing Jin. 2023.
\newblock \href {https://doi.org/10.1038/s42256-023-00729-y} {{A taxonomy and review of generalization research in NLP}}.
\newblock \emph{Nature Machine Intelligence}, 5(10):1161--1174.

\bibitem[{Ji et~al.(2024)Ji, Li, Paul, Paavola, Lin, Chen, O'Brien, Luo, Sch{\"u}tze, Tiedemann, and Haddow}]{jiEMMA500EnhancingMassively2024}
Shaoxiong Ji, Zihao Li, Indraneil Paul, Jaakko Paavola, Peiqin Lin, Pinzhen Chen, Dayy{\'a}n O'Brien, Hengyu Luo, Hinrich Sch{\"u}tze, J{\"o}rg Tiedemann, and Barry Haddow. 2024.
\newblock \href {https://doi.org/10.48550/arXiv.2409.17892} {{{EMMA-500}}: {{Enhancing Massively Multilingual Adaptation}} of {{Large Language Models}}}.
\newblock \emph{Preprint}, arXiv:2409.17892.

\bibitem[{Jumelet et~al.(2021)Jumelet, Denic, Szymanik, Hupkes, and Steinert-Threlkeld}]{jumelet-etal-2021-language}
Jaap Jumelet, Milica Denic, Jakub Szymanik, Dieuwke Hupkes, and Shane Steinert-Threlkeld. 2021.
\newblock \href {https://doi.org/10.18653/v1/2021.findings-acl.439} {Language models use monotonicity to assess {NPI} licensing}.
\newblock In \emph{Findings of the Association for Computational Linguistics: ACL-IJCNLP 2021}, pages 4958--4969, Online. Association for Computational Linguistics.

\bibitem[{Jurafsky and Martin(2025)}]{jm3nlp}
Daniel Jurafsky and James~H. Martin. 2025.
\newblock \href {https://web.stanford.edu/~jurafsky/slp3/24.pdf} {Discourse coherence}.
\newblock In \emph{Speech and Language Processing: An Introduction to Natural Language Processing, Computational Linguistics, and Speech Recognition with Language Models}, 3rd edition, chapter~24, pages 531--552.
\newblock Online manuscript released January 12, 2025.

\bibitem[{Kim and Schuster(2023)}]{kim-schuster-2023-entity}
Najoung Kim and Sebastian Schuster. 2023.
\newblock \href {https://doi.org/10.18653/v1/2023.acl-long.213} {Entity tracking in language models}.
\newblock In \emph{Proceedings of the 61st Annual Meeting of the Association for Computational Linguistics (Volume 1: Long Papers)}, pages 3835--3855, Toronto, Canada. Association for Computational Linguistics.

\bibitem[{Knaebel(2021)}]{knaebel-2021-discopy}
Ren{\'e} Knaebel. 2021.
\newblock \href {https://doi.org/10.18653/v1/2021.codi-main.12} {discopy: A neural system for shallow discourse parsing}.
\newblock In \emph{Proceedings of the 2nd Workshop on Computational Approaches to Discourse}, pages 128--133, Punta Cana, Dominican Republic and Online. Association for Computational Linguistics.

\bibitem[{Koto et~al.(2021)Koto, Lau, and Baldwin}]{koto-etal-2021-discourse}
Fajri Koto, Jey~Han Lau, and Timothy Baldwin. 2021.
\newblock \href {https://doi.org/10.18653/v1/2021.naacl-main.301} {Discourse probing of pretrained language models}.
\newblock In \emph{Proceedings of the 2021 Conference of the North American Chapter of the Association for Computational Linguistics: Human Language Technologies}, pages 3849--3864, Online. Association for Computational Linguistics.

\bibitem[{Kurfal{\i} and {\"O}stling(2021)}]{kurfali-ostling-2021-probing}
Murathan Kurfal{\i} and Robert {\"O}stling. 2021.
\newblock \href {https://doi.org/10.18653/v1/2021.repl4nlp-1.2} {Probing multilingual language models for discourse}.
\newblock In \emph{Proceedings of the 6th Workshop on Representation Learning for NLP (RepL4NLP-2021)}, pages 8--19, Online. Association for Computational Linguistics.

\bibitem[{Li et~al.(2020)Li, Liu, Kan, Zheng, Wang, Lei, Liu, and Qin}]{li-etal-2020-molweni}
Jiaqi Li, Ming Liu, Min-Yen Kan, Zihao Zheng, Zekun Wang, Wenqiang Lei, Ting Liu, and Bing Qin. 2020.
\newblock \href {https://doi.org/10.18653/v1/2020.coling-main.238} {Molweni: A challenge multiparty dialogues-based machine reading comprehension dataset with discourse structure}.
\newblock In \emph{Proceedings of the 28th International Conference on Computational Linguistics}, pages 2642--2652, Barcelona, Spain (Online). International Committee on Computational Linguistics.

\bibitem[{Li et~al.(2014)Li, Wang, Cao, and Li}]{li-etal-2014-text}
Sujian Li, Liang Wang, Ziqiang Cao, and Wenjie Li. 2014.
\newblock \href {https://doi.org/10.3115/v1/P14-1003} {Text-level discourse dependency parsing}.
\newblock In \emph{Proceedings of the 52nd Annual Meeting of the Association for Computational Linguistics (Volume 1: Long Papers)}, pages 25--35, Baltimore, Maryland. Association for Computational Linguistics.

\bibitem[{Liu and Zeldes(2023)}]{liu-zeldes-2023-cant}
Yang~Janet Liu and Amir Zeldes. 2023.
\newblock \href {https://doi.org/10.18653/v1/2023.eacl-main.227} {Why can`t discourse parsing generalize? a thorough investigation of the impact of data diversity}.
\newblock In \emph{Proceedings of the 17th Conference of the European Chapter of the Association for Computational Linguistics}, pages 3112--3130, Dubrovnik, Croatia. Association for Computational Linguistics.

\bibitem[{Liu et~al.(2021)Liu, Shi, and Chen}]{liu-etal-2021-dmrst}
Zhengyuan Liu, Ke~Shi, and Nancy Chen. 2021.
\newblock \href {https://doi.org/10.18653/v1/2021.codi-main.15} {{DMRST}: A joint framework for document-level multilingual {RST} discourse segmentation and parsing}.
\newblock In \emph{Proceedings of the 2nd Workshop on Computational Approaches to Discourse}, pages 154--164, Punta Cana, Dominican Republic and Online. Association for Computational Linguistics.

\bibitem[{Loshchilov and Hutter(2018)}]{loshchilovDecoupledWeightDecay2018}
Ilya Loshchilov and Frank Hutter. 2018.
\newblock \href {https://openreview.net/forum?id=Bkg6RiCqY7} {Decoupled {{Weight Decay Regularization}}}.
\newblock In \emph{International {{Conference}} on {{Learning Representations}}}.

\bibitem[{Mann and Thompson(1988)}]{mann1988rhetorical}
William~C Mann and Sandra~A Thompson. 1988.
\newblock Rhetorical {S}tructure {T}heory: Toward a {F}unctional {T}heory of {T}ext {O}rganization.
\newblock \emph{Text-Interdisciplinary Journal for the Study of Discourse}, 8(3):243--281.

\bibitem[{Miao et~al.(2024)Miao, Liu, Lei, Chen, and Kan}]{miao-etal-2024-discursive}
Yisong Miao, Hongfu Liu, Wenqiang Lei, Nancy Chen, and Min-Yen Kan. 2024.
\newblock \href {https://doi.org/10.18653/v1/2024.acl-long.341} {Discursive socratic questioning: Evaluating the faithfulness of language models' understanding of discourse relations}.
\newblock In \emph{Proceedings of the 62nd Annual Meeting of the Association for Computational Linguistics (Volume 1: Long Papers)}, pages 6277--6295, Bangkok, Thailand. Association for Computational Linguistics.

\bibitem[{Mistral(2025)}]{mistralMistralSmall32025}
AI~Team Mistral. 2025.
\newblock Mistral {{Small}} 3.
\newblock https://mistral.ai/en/news/mistral-small-3.

\bibitem[{Morey et~al.(2018)Morey, Muller, and Asher}]{morey-etal-2018-dependency}
Mathieu Morey, Philippe Muller, and Nicholas Asher. 2018.
\newblock \href {https://doi.org/10.1162/COLI_a_00314} {A dependency perspective on {RST} discourse parsing and evaluation}.
\newblock \emph{Computational Linguistics}, 44(2):197--235.

\bibitem[{Mrini et~al.(2020)Mrini, Dernoncourt, Tran, Bui, Chang, and Nakashole}]{mrini-etal-2020-rethinking}
Khalil Mrini, Franck Dernoncourt, Quan~Hung Tran, Trung Bui, Walter Chang, and Ndapa Nakashole. 2020.
\newblock \href {https://doi.org/10.18653/v1/2020.findings-emnlp.65} {Rethinking self-attention: Towards interpretability in neural parsing}.
\newblock In \emph{Findings of the Association for Computational Linguistics: EMNLP 2020}, pages 731--742, Online. Association for Computational Linguistics.

\bibitem[{Muller et~al.(2012)Muller, Afantenos, Denis, and Asher}]{muller-etal-2012-constrained}
Philippe Muller, Stergos Afantenos, Pascal Denis, and Nicholas Asher. 2012.
\newblock \href {https://aclanthology.org/C12-1115/} {Constrained decoding for text-level discourse parsing}.
\newblock In \emph{Proceedings of {COLING} 2012}, pages 1883--1900, Mumbai, India. The COLING 2012 Organizing Committee.

\bibitem[{Narasimhan and Barzilay(2015)}]{narasimhan-barzilay-2015-machine}
Karthik Narasimhan and Regina Barzilay. 2015.
\newblock \href {https://doi.org/10.3115/v1/P15-1121} {Machine comprehension with discourse relations}.
\newblock In \emph{Proceedings of the 53rd Annual Meeting of the Association for Computational Linguistics and the 7th International Joint Conference on Natural Language Processing (Volume 1: Long Papers)}, pages 1253--1262, Beijing, China. Association for Computational Linguistics.

\bibitem[{Peng and S{\o}gaard(2024)}]{peng-sogaard-2024-concept}
Qiwei Peng and Anders S{\o}gaard. 2024.
\newblock \href {https://doi.org/10.18653/v1/2024.emnlp-main.315} {Concept space alignment in multilingual {LLM}s}.
\newblock In \emph{Proceedings of the 2024 Conference on Empirical Methods in Natural Language Processing}, pages 5511--5526, Miami, Florida, USA. Association for Computational Linguistics.

\bibitem[{Pimentel et~al.(2021)Pimentel, Ryskina, Mielke, Wu, Chodroff, Leonard, Nicolai, Ghanggo~Ate, Khalifa, Habash, El-Khaissi, Goldman, Gasser, Lane, Coler, Oncevay, Montoya~Samame, Silva~Villegas, Ek, Bernardy, Shcherbakov, Bayyr-ool, Sheifer, Ganieva, Plugaryov, Klyachko, Salehi, Krizhanovsky, Krizhanovsky, Vania, Ivanova, Salchak, Straughn, Liu, Washington, Ataman, Kiera{\'s}, Woli{\'n}ski, Suhardijanto, Stoehr, Nuriah, Ratan, Tyers, Ponti, Aiton, Hatcher, Prud{'}hommeaux, Kumar, Hulden, Barta, Lakatos, Szolnok, {\'A}cs, Raj, Yarowsky, Cotterell, Ambridge, and Vylomova}]{pimentel-ryskina-etal-2021-sigmorphon}
Tiago Pimentel, Maria Ryskina, Sabrina~J. Mielke, Shijie Wu, Eleanor Chodroff, Brian Leonard, Garrett Nicolai, Yustinus Ghanggo~Ate, Salam Khalifa, Nizar Habash, Charbel El-Khaissi, Omer Goldman, Michael Gasser, William Lane, Matt Coler, Arturo Oncevay, Jaime~Rafael Montoya~Samame, Gema~Celeste Silva~Villegas, Adam Ek, and 39 others. 2021.
\newblock \href {https://doi.org/10.18653/v1/2021.sigmorphon-1.25} {{SIGMORPHON} 2021 shared task on morphological reinflection: Generalization across languages}.
\newblock In \emph{Proceedings of the 18th SIGMORPHON Workshop on Computational Research in Phonetics, Phonology, and Morphology}, pages 229--259, Online. Association for Computational Linguistics.

\bibitem[{Poesio et~al.(2024)Poesio, Ogrodniczuk, Ng, Pradhan, Yu, Moosavi, Paun, Zeldes, Nedoluzhko, Nov{\'a}k, Popel, {\v{Z}}abokrtsk{\'y}, and Zeman}]{poesio-etal-2024-universal}
Massimo Poesio, Maciej Ogrodniczuk, Vincent Ng, Sameer Pradhan, Juntao Yu, Nafise~Sadat Moosavi, Silviu Paun, Amir Zeldes, Anna Nedoluzhko, Michal Nov{\'a}k, Martin Popel, Zden{\v{e}}k {\v{Z}}abokrtsk{\'y}, and Daniel Zeman. 2024.
\newblock \href {https://aclanthology.org/2024.lrec-main.1484/} {Universal anaphora: The first three years}.
\newblock In \emph{Proceedings of the 2024 Joint International Conference on Computational Linguistics, Language Resources and Evaluation (LREC-COLING 2024)}, pages 17087--17100, Torino, Italia. ELRA and ICCL.

\bibitem[{Prasertsom et~al.(2024)Prasertsom, Jaroonpol, and Rutherford}]{ThaiPDTB}
Ponrawee Prasertsom, Apiwat Jaroonpol, and Attapol~T. Rutherford. 2024.
\newblock \href {https://doi.org/10.1162/tacl_a_00650} {{The Thai Discourse Treebank: Annotating and Classifying Thai Discourse Connectives}}.
\newblock \emph{Transactions of the Association for Computational Linguistics}, 12:613--629.

\bibitem[{Qwen et~al.(2025)Qwen, Yang, Yang, Zhang, Hui, Zheng, Yu, Li, Liu, Huang, Wei, Lin, Yang, Tu, Zhang, Yang, Yang, Zhou, Lin, Dang, Lu, Bao, Yang, Yu, Li, Xue, Zhang, Zhu, Men, Lin, Li, Tang, Xia, Ren, Ren, Fan, Su, Zhang, Wan, Liu, Cui, Zhang, and Qiu}]{qwenQwen25TechnicalReport2025}
Qwen, An~Yang, Baosong Yang, Beichen Zhang, Binyuan Hui, Bo~Zheng, Bowen Yu, Chengyuan Li, Dayiheng Liu, Fei Huang, Haoran Wei, Huan Lin, Jian Yang, Jianhong Tu, Jianwei Zhang, Jianxin Yang, Jiaxi Yang, Jingren Zhou, Junyang Lin, and 24 others. 2025.
\newblock \href {https://doi.org/10.48550/arXiv.2412.15115} {Qwen2.5 {{Technical Report}}}.
\newblock \emph{Preprint}, arXiv:2412.15115.

\bibitem[{Rehbein et~al.(2016)Rehbein, Scholman, and Demberg}]{rehbein-etal-2016-annotating}
Ines Rehbein, Merel Scholman, and Vera Demberg. 2016.
\newblock \href {https://aclanthology.org/L16-1165/} {Annotating discourse relations in spoken language: A comparison of the {PDTB} and {CCR} frameworks}.
\newblock In \emph{Proceedings of the Tenth International Conference on Language Resources and Evaluation ({LREC}`16)}, pages 1039--1046, Portoro{\v{z}}, Slovenia. European Language Resources Association (ELRA).

\bibitem[{Sanders et~al.(1992)Sanders, Spooren, and Noordman}]{SandersSpoorenNoordman1992}
Ted J.~M. Sanders, Wilbert~P.M. Spooren, and Leo~G.M. Noordman. 1992.
\newblock Towards a taxonomy of coherence relations.
\newblock \emph{Discourse Processes}, 15:1--35.

\bibitem[{Sanders et~al.(2021)Sanders, Demberg, Hoek, Scholman, Asr, Zufferey, and Evers-Vermeul}]{SandersDembergHoekScholmanAsrZuffereyEversVermeul}
Ted~J.M. Sanders, Vera Demberg, Jet Hoek, Merel~C.J. Scholman, Fatemeh~Torabi Asr, Sandrine Zufferey, and Jacqueline Evers-Vermeul. 2021.
\newblock \href {https://doi.org/doi:10.1515/cllt-2016-0078} {Unifying dimensions in coherence relations: How various annotation frameworks are related}.
\newblock \emph{Corpus Linguistics and Linguistic Theory}, 17(1):1--71.

\bibitem[{Saputa et~al.(2024)Saputa, Peljak-{\L}api{\'n}ska, and Ogrodniczuk}]{saputa-etal-2024-polish}
Karol Saputa, Angelika Peljak-{\L}api{\'n}ska, and Maciej Ogrodniczuk. 2024.
\newblock \href {https://doi.org/10.18653/v1/2024.crac-1.3} {{P}olish coreference corpus as an {LLM} testbed: Evaluating coreference resolution within instruction-following language models by instruction{--}answer alignment}.
\newblock In \emph{Proceedings of The Seventh Workshop on Computational Models of Reference, Anaphora and Coreference}, pages 23--32, Miami. Association for Computational Linguistics.

\bibitem[{Scao et~al.(2023)Scao, Fan, Akiki, Pavlick, Ili{\'c}, Hesslow, Castagn{\'e}, Luccioni, Yvon, Gall{\'e}, Tow, Rush, Biderman, Webson, Ammanamanchi, Wang, Sagot, Muennighoff, del Moral, Ruwase, Bawden, Bekman, {McMillan-Major}, Beltagy, Nguyen, Saulnier, Tan, Suarez, Sanh, Lauren{\c c}on, Jernite, Launay, Mitchell, Raffel, Gokaslan, Simhi, Soroa, Aji, Alfassy, Rogers, Nitzav, Xu, Mou, Emezue, Klamm, Leong, van Strien, Adelani, Radev, Ponferrada, Levkovizh, Kim, Natan, Toni, Dupont, Kruszewski, Pistilli, Elsahar, Benyamina, Tran, Yu, Abdulmumin, Johnson, {Gonzalez-Dios}, de~la Rosa, Chim, Dodge, Zhu, Chang, Frohberg, Tobing, Bhattacharjee, Almubarak, Chen, Lo, Werra, Weber, Phan, {allal}, Tanguy, Dey, Mu{\~n}oz, Masoud, Grandury, {\v S}a{\v s}ko, Huang, Coavoux, Singh, Jiang, Vu, Jauhar, Ghaleb, Subramani, Kassner, Khamis, Nguyen, Espejel, de~Gibert, Villegas, Henderson, Colombo, Amuok, Lhoest, Harliman, Bommasani, L{\'o}pez, Ribeiro, Osei, Pyysalo, Nagel, Bose, Muhammad, Sharma, Longpre, Nikpoor,
  Silberberg, Pai, Zink, Torrent, Schick, Thrush, Danchev, Nikoulina, Laippala, Lepercq, Prabhu, Alyafeai, Talat, Raja, Heinzerling, Si, Ta{\c s}ar, Salesky, Mielke, Lee, Sharma, Santilli, Chaffin, Stiegler, Datta, Szczechla, Chhablani, Wang, Pandey, Strobelt, Fries, Rozen, Gao, Sutawika, Bari, {Al-shaibani}, Manica, Nayak, Teehan, Albanie, Shen, {Ben-David}, Bach, Kim, Bers, Fevry, Neeraj, Thakker, Raunak, Tang, Yong, Sun, Brody, Uri, Tojarieh, Roberts, Chung, Tae, Phang, Press, Li, Narayanan, Bourfoune, Casper, Rasley, Ryabinin, Mishra, Zhang, Shoeybi, Peyrounette, Patry, Tazi, Sanseviero, von Platen, Cornette, Lavall{\'e}e, Lacroix, Rajbhandari, Gandhi, Smith, Requena, Patil, Dettmers, Baruwa, Singh, Cheveleva, Ligozat, Subramonian, N{\'e}v{\'e}ol, Lovering, Garrette, Tunuguntla, Reiter, Taktasheva, Voloshina, Bogdanov, Winata, Schoelkopf, Kalo, Novikova, Forde, Clive, Kasai, Kawamura, Hazan, Carpuat, Clinciu, Kim, Cheng, Serikov, Antverg, van~der Wal, Zhang, Zhang, Gehrmann, Mirkin, Pais, Shavrina,
  Scialom, Yun, Limisiewicz, Rieser, Protasov, Mikhailov, Pruksachatkun, Belinkov, Bamberger, Kasner, Rueda, Pestana, Feizpour, Khan, Faranak, Santos, Hevia, Unldreaj, Aghagol, Abdollahi, Tammour, HajiHosseini, Behroozi, Ajibade, Saxena, Ferrandis, McDuff, Contractor, Lansky, David, Kiela, Nguyen, Tan, Baylor, Ozoani, Mirza, Ononiwu, Rezanejad, Jones, Bhattacharya, Solaiman, Sedenko, Nejadgholi, Passmore, Seltzer, Sanz, Dutra, Samagaio, Elbadri, Mieskes, Gerchick, Akinlolu, McKenna, Qiu, Ghauri, Burynok, Abrar, Rajani, Elkott, Fahmy, Samuel, An, Kromann, Hao, Alizadeh, Shubber, Wang, Roy, Viguier, Le, Oyebade, Le, Yang, Nguyen, Kashyap, Palasciano, Callahan, Shukla, {Miranda-Escalada}, Singh, Beilharz, Wang, Brito, Zhou, Jain, Xu, Fourrier, Peri{\~n}{\'a}n, Molano, Yu, Manjavacas, Barth, Fuhrimann, Altay, Bayrak, Burns, Vrabec, Bello, Dash, Kang, Giorgi, Golde, Posada, Sivaraman, Bulchandani, Liu, Shinzato, de~Bykhovetz, Takeuchi, P{\`a}mies, Castillo, Nezhurina, S{\"a}nger, Samwald, Cullan, Weinberg, Wolf,
  Mihaljcic, Liu, Freidank, Kang, Seelam, Dahlberg, Broad, Muellner, Fung, Haller, Chandrasekhar, Eisenberg, Martin, Canalli, Su, Su, Cahyawijaya, Garda, Deshmukh, Mishra, Kiblawi, Ott, {Sang-aroonsiri}, Kumar, Schweter, Bharati, Laud, Gigant, Kainuma, Kusa, Labrak, Bajaj, Venkatraman, Xu, Xu, Xu, Tan, Xie, Ye, Bras, Belkada, and Wolf}]{scaoBLOOM176BParameterOpenAccess2023}
Teven~Le Scao, Angela Fan, Christopher Akiki, Ellie Pavlick, Suzana Ili{\'c}, Daniel Hesslow, Roman Castagn{\'e}, Alexandra~Sasha Luccioni, Fran{\c c}ois Yvon, Matthias Gall{\'e}, Jonathan Tow, Alexander~M. Rush, Stella Biderman, Albert Webson, Pawan~Sasanka Ammanamanchi, Thomas Wang, Beno{\^i}t Sagot, Niklas Muennighoff, Albert~Villanova del Moral, and 373 others. 2023.
\newblock \href {https://doi.org/10.48550/arXiv.2211.05100} {{{BLOOM}}: {{A 176B-Parameter Open-Access Multilingual Language Model}}}.
\newblock \emph{Preprint}, arXiv:2211.05100.

\bibitem[{Schiffrin et~al.(2015)Schiffrin, Tannen, and E.~Hamilton}]{SchiffrinEtAl2015Ch0}
Deborah Schiffrin, Deborah Tannen, and Heidi E.~Hamilton. 2015.
\newblock \href {https://doi.org/10.1002/9781118584194.ch0} {Introduction to the first edition}.
\newblock In \emph{The Handbook of Discourse Analysis}, chapter~00, pages 1--7. John Wiley \& Sons, Ltd.

\bibitem[{Scivetti et~al.(2025)Scivetti, Torgbi, Blodgett, Shichman, Hudson, Bonial, and Madabushi}]{scivetti2025assessinglanguagecomprehensionlarge}
Wesley Scivetti, Melissa Torgbi, Austin Blodgett, Mollie Shichman, Taylor Hudson, Claire Bonial, and Harish~Tayyar Madabushi. 2025.
\newblock \href {https://arxiv.org/abs/2501.04661} {Assessing language comprehension in large language models using construction grammar}.
\newblock \emph{Preprint}, arXiv:2501.04661.

\bibitem[{Skean et~al.(2025)Skean, Arefin, Zhao, Patel, Naghiyev, LeCun, and Shwartz{-}Ziv}]{skean2025layerlayeruncoveringhidden}
Oscar Skean, Md~Rifat Arefin, Dan Zhao, Niket Patel, Jalal Naghiyev, Yann LeCun, and Ravid Shwartz{-}Ziv. 2025.
\newblock \href {https://doi.org/10.48550/ARXIV.2502.02013} {{Layer by Layer: Uncovering Hidden Representations in Language Models}}.
\newblock \emph{CoRR}, abs/2502.02013.

\bibitem[{Stede(2011)}]{stede2011discourse}
Manfred Stede. 2011.
\newblock Discourse {P}rocessing.
\newblock \emph{Synthesis Lectures on Human Language Technologies}, 4(3):1--165.

\bibitem[{Tenney et~al.(2019)Tenney, Xia, Chen, Wang, Poliak, McCoy, Kim, Durme, Bowman, Das, and Pavlick}]{tenneyWhatYouLearn2019}
Ian Tenney, Patrick Xia, Berlin Chen, Alex Wang, Adam Poliak, R~Thomas McCoy, Najoung Kim, Benjamin~Van Durme, Sam Bowman, Dipanjan Das, and Ellie Pavlick. 2019.
\newblock \href {https://openreview.net/forum?id=SJzSgnRcKX} {What do you learn from context? probing for sentence structure in contextualized word representations}.
\newblock In \emph{International Conference on Learning Representations}.

\bibitem[{Toraman et~al.(2023)Toraman, Yilmaz, \c{S}ahinuc, and Ozcelik}]{2022tokenizerImpact}
Cagri Toraman, Eyup~Halit Yilmaz, Furkan \c{S}ahinuc, and Oguzhan Ozcelik. 2023.
\newblock \href {https://doi.org/10.1145/3578707} {Impact of tokenization on language models: An analysis for turkish}.
\newblock \emph{ACM Trans. Asian Low-Resour. Lang. Inf. Process.}, 22(4).

\bibitem[{Vaswani et~al.(2017)Vaswani, Shazeer, Parmar, Uszkoreit, Jones, Gomez, Kaiser, and Polosukhin}]{2017nipsAttentionIsAllYouNeed}
Ashish Vaswani, Noam Shazeer, Niki Parmar, Jakob Uszkoreit, Llion Jones, Aidan~N Gomez, \L~ukasz Kaiser, and Illia Polosukhin. 2017.
\newblock \href {https://proceedings.neurips.cc/paper_files/paper/2017/file/3f5ee243547dee91fbd053c1c4a845aa-Paper.pdf} {Attention is all you need}.
\newblock In \emph{Advances in Neural Information Processing Systems}, volume~30. Curran Associates, Inc.

\bibitem[{Webber et~al.(2024)Webber, Prasad, and Josh}]{WebberEtAl2024PDTB}
Bonnie Webber, Rashmi Prasad, and Aravind Josh. 2024.
\newblock \href {https://submissions.cljournal.org/index.php/cljournal/article/view/543} {{Reflections on the Penn Discourse TreeBank and its relatives}}.
\newblock \emph{Computational Linguistics}.

\bibitem[{Webber et~al.(2019)Webber, Prasad, Lee, and Joshi}]{webber2019penn}
Bonnie Webber, Rashmi Prasad, Alan Lee, and Aravind Joshi. 2019.
\newblock \href {https://catalog.ldc.upenn.edu/docs/LDC2019T05/PDTB3-Annotation-Manual.pdf} {{The Penn Discourse Treebank 3.0 Annotation Manual}}.
\newblock \emph{Philadelphia, University of Pennsylvania}.

\bibitem[{Weissweiler et~al.(2022)Weissweiler, Hofmann, K{\"o}ksal, and Sch{\"u}tze}]{weissweiler-etal-2022-better}
Leonie Weissweiler, Valentin Hofmann, Abdullatif K{\"o}ksal, and Hinrich Sch{\"u}tze. 2022.
\newblock \href {https://doi.org/10.18653/v1/2022.emnlp-main.746} {The better your syntax, the better your semantics? probing pretrained language models for the {E}nglish comparative correlative}.
\newblock In \emph{Proceedings of the 2022 Conference on Empirical Methods in Natural Language Processing}, pages 10859--10882, Abu Dhabi, United Arab Emirates. Association for Computational Linguistics.

\bibitem[{Wendler et~al.(2024)Wendler, Veselovsky, Monea, and West}]{wendler-etal-2024-llamas}
Chris Wendler, Veniamin Veselovsky, Giovanni Monea, and Robert West. 2024.
\newblock \href {https://doi.org/10.18653/v1/2024.acl-long.820} {{Do Llamas Work in {E}nglish? On the Latent Language of Multilingual Transformers}}.
\newblock In \emph{Proceedings of the 62nd Annual Meeting of the Association for Computational Linguistics (Volume 1: Long Papers)}, pages 15366--15394, Bangkok, Thailand. Association for Computational Linguistics.

\bibitem[{Xu et~al.(2020)Xu, Gan, Cheng, and Liu}]{xu-etal-2020-discourse}
Jiacheng Xu, Zhe Gan, Yu~Cheng, and Jingjing Liu. 2020.
\newblock \href {https://doi.org/10.18653/v1/2020.acl-main.451} {Discourse-aware neural extractive text summarization}.
\newblock In \emph{Proceedings of the 58th Annual Meeting of the Association for Computational Linguistics}, pages 5021--5031, Online. Association for Computational Linguistics.

\bibitem[{Zeldes et~al.(2025)Zeldes, Aoyama, Liu, Peng, Das, and Gessler}]{ZeldesEtAl2024erst}
Amir Zeldes, Tatsuya Aoyama, Yang~Janet Liu, Siyao Peng, Debopam Das, and Luke Gessler. 2025.
\newblock \href {https://doi.org/10.1162/coli_a_00538} {{eRST: A Signaled Graph Theory of Discourse Relations and Organization}}.
\newblock \emph{Computational Linguistics}, 51(1):23--72.

\bibitem[{Zeldes et~al.(2021)Zeldes, Liu, Iruskieta, Muller, Braud, and Badene}]{zeldes-etal-2021-disrpt}
Amir Zeldes, Yang~Janet Liu, Mikel Iruskieta, Philippe Muller, Chlo{\'e} Braud, and Sonia Badene. 2021.
\newblock \href {https://doi.org/10.18653/v1/2021.disrpt-1.1} {The {DISRPT} 2021 shared task on elementary discourse unit segmentation, connective detection, and relation classification}.
\newblock In \emph{Proceedings of the 2nd Shared Task on Discourse Relation Parsing and Treebanking (DISRPT 2021)}, pages 1--12, Punta Cana, Dominican Republic. Association for Computational Linguistics.

\end{thebibliography}

\clearpage
\newpage
\appendix

\section{Unified Label Set: Definitions and Examples}
\label{app:labels}

The proposed unified label set adapts the mapping proposal described in \citet{benamara-taboada-2015-mapping} and extends it to be applicable to phenomena frequent in dialogues. In total, there are $17$ labels corresponding to the core discourse relations identified in \citet{Bunt2016ISOD}.  
Below we provide definitions and list typical framework-specific discourse relations that are mapped to the unified label set. 

\paragraph{\textsc{temporal}}\!is used to map to framework-specific labels that establish a chronological sequence between events or states. RST's \textsc{sequence}, PDTB's \textsc{temporal.asynchronous}/\textsc{synchronous}, and SDRT's \textsc{temploc} and \textsc{flashback} are mapped to this class. 

\paragraph{\textsc{structuring}}\!is mapped to RST-style relations such as \textsc{list}, \textsc{preparation}, \textsc{disjunction}, \textsc{organization-heading}, \textsc{joint}, and \textsc{textual-organization}. SDRT-style relations such as \textsc{alternation}, \textsc{continuation}, and \textsc{parallel} are mapped to this class. PDTB's \textsc{expansion.disjunction} is mapped to this class. 

\paragraph{\textsc{attribution}}\!informs about the source of information, which is useful and crucial for many real-world applications such as misinformation detection and fact-checking. It is considered a discourse relation by RST, SDRT, and DEP. PDTB's attribution annotation is considered a separate discourse annotation type and is not included in DISRPT. It is worth pointing out that \textsc{attribution} is not considered a core discourse relation by \citet{Bunt2016ISOD}, but it is one of the most frequent discourse relations in most frameworks. 

\paragraph{\textsc{comparison}}\!is used to group fine-grained relations that highlight similarities rather than differences between spans or entities such as PDTB's \textsc{comparison.similarity} and \textsc{analogy} in RST-style corpora. 

\paragraph{\textsc{elaboration}}\!provides additional information about an entity or a proposition. Framework-specific relations that contain \textsc{elaboration} or are prefixed with \textsc{elab} (e.g.~\textsc{elab-process\_step}, \textsc{elab-enumember}, and \textsc{Q\_Elab}) are all mapped to this class. \textsc{progression} used in DEP-style corpora is also mapped to this class.  

\paragraph{\textsc{framing}} is used for framework-specific relations that provide a framework for understanding the content of the situation described in the discourse units \cite{benamara-taboada-2015-mapping} such as \textsc{frame}, \textsc{background}, and \textsc{circumstance}

\paragraph{\textsc{mode}}\! is used to supply information about \textit{how} events happens. Commonly mapped fine-grained relations are \textsc{manner} and \textsc{means} in RST-style corpora as well as PDTB's \textsc{expansion.manner}.

\paragraph{\textsc{reformulation}}\!corresponds to relations by which one discourse unit re-expresses the meaning of another in a different form and ensure coherence by providing alternative expressions of the same idea. \textsc{summary} and \textsc{restatement} in RST-style corpora and PDTB's \textsc{expansion.equivalence} are mapped to \textsc{reformulation}. 

\paragraph{\textsc{adversative}}\!highlights incompatibility and covers commonly used discourse relations in all frameworks such as \textsc{concession} and \textsc{contrast}. PDTB's \textsc{expansion.exception}/\textsc{Substitution} are also mapped to this subclass given their definitions in \citet{webber2019penn}. 

\paragraph{\textsc{causal}}\!is used to indicate a cause-and-effect relationship. Fine-grained relations that signal that one event, state, or proposition (the cause) leads to or explains another event, state, or proposition (the effect) is mapped to this class. This is one of the most core discourse relation types recognized in all frameworks. 

\paragraph{\textsc{contingency}}\!is used to map condition-based relations such as \textsc{conditional}, \textsc{unless}, \textsc{unconditional}, and \textsc{contingency.negative-condition}. 

\paragraph{\textsc{enablement}}\!is used to connect discourse units where one enables the other. Framework-specific relations such as \textsc{goal} and \textsc{purpose} are mapped to this class. 

\paragraph{\textsc{explanation}}\!is used when the situation described by one argument provides the reason, explanation, or justification for the situation described by the other \cite{webber2019penn}.

\paragraph{\textsc{evaluation}}\!is used where one discourse unit provides an assessment, judgment, or commentary on the content of another unit. Framework-dependent relations such as \textsc{comment} and \textsc{interpretation} are mapped to this class.

\paragraph{\textsc{topic-change}}\!involves a shift or drift in topic that links large textual units \cite{carlson-marcu-01}. This is used for fine-grained relations that connect multiple, non-contrasting discourse units that are of equal prominence such as \textsc{joint-other} in the \texttt{eng.rst.gum} corpus and PDTB's \textsc{expansion.conjunction}. 

\paragraph{\textsc{topic-comment}}\!is used for framework-specific relatiosn that involve question-answer pairs or problem-solution pairs, commonly in RST-style and SDRT-style corpora. PDTB's \textsc{hypophora} is also mapped to this class.

\paragraph{\textsc{topic-adjustment}}\!is primarily used for cases where a discourse unit modifies, redirects, or adjusts the ongoing topic of discussion such as \textsc{correction} and \textsc{interrupted}, which signal deviations from the expected discourse progression.

\section{Data}
\label{app:data-disrpt}

\autoref{tab:disrpt-overview} provides an overview of the DISRPT benchmark we use for our experiments. The data produced and used in this paper is in accordance with the original licenses of the underlying resources, as specified in the repository of DISRPT.\!\footnote{\url{https://github.com/disrpt/sharedtask2023}} 

\begin{table}[ht]
\centering
\resizebox{\columnwidth}{!}{%
\begin{tabular}{@{}l|l|l|r|r@{}}
\toprule
\multicolumn{1}{c|}{\textbf{dataset}} & \multicolumn{1}{c|}{\textbf{language}} & \multicolumn{1}{c|}{\textbf{\begin{tabular}[c]{@{}c@{}}language \\ family\end{tabular}}} & \multicolumn{1}{c|}{\textbf{\begin{tabular}[c]{@{}c@{}}\# of relation\\ instances\end{tabular}}} & \multicolumn{1}{l}{\textbf{\begin{tabular}[c]{@{}l@{}}\% of total \\ instances\end{tabular}}} \\ \midrule
deu.rst.pcc & German & \multirow{8}{*}{\begin{tabular}[c]{@{}l@{}}Indo-European, \\ Germanic\end{tabular}} & 2665 & 1.19\% \\ \cmidrule(r){1-2} \cmidrule(l){4-5} 
eng.dep.covdtb & \multirow{7}{*}{English} &  & 4985 & 2.22\% \\
eng.dep.scidtb &  &  & 9904 & 4.42\% \\
eng.pdtb.pdtb &  &  & 47851 & 21.34\% \\
eng.pdtb.tedm &  &  & 529 & 0.24\% \\
eng.rst.gum &  &  & 24688 & 11.01\% \\
eng.rst.rstdt &  &  & 19778 & 8.82\% \\
eng.sdrt.stac &  &  & 12235 & 5.46\% \\ \midrule
eus.rst.ert & Basque & \begin{tabular}[c]{@{}l@{}}Language \\ Isolate\end{tabular} & 3825 & 1.71\% \\ \midrule
fas.rst.prstc & Farsi & \begin{tabular}[c]{@{}l@{}}Indo-European, \\ Iranian\end{tabular} & 5191 & 2.31\% \\ \midrule
fra.sdrt.annodis & French & \multirow{2}{*}{\begin{tabular}[c]{@{}l@{}}Indo-European, \\ Romance\end{tabular}} & 3338 & 1.49\% \\ \cmidrule(r){1-2} \cmidrule(l){4-5} 
ita.pdtb.luna & Italian &  & 1544 & 0.69\% \\ \midrule
nld.rst.nldt & Dutch & \begin{tabular}[c]{@{}l@{}}Indo-European, \\ Germanic\end{tabular} & 2264 & 1.01\% \\ \midrule
por.pdtb.crpc & \multirow{3}{*}{Portuguese} & \multirow{3}{*}{\begin{tabular}[c]{@{}l@{}}Indo-European, \\ Romance\end{tabular}} & 11330 & 5.05\% \\
por.pdtb.tedm &  &  & 554 & 0.25\% \\
por.rst.cstn &  &  & 4993 & 2.23\% \\ \midrule
rus.rst.rrt & Russian & \begin{tabular}[c]{@{}l@{}}Indo-European, \\ Slavic\end{tabular} & 34566 & 15.41\% \\ \midrule
spa.rst.rststb & \multirow{2}{*}{Spanish} & \multirow{2}{*}{\begin{tabular}[c]{@{}l@{}}Indo-European, \\ Romance\end{tabular}} & 3049 & 1.36\% \\
spa.rst.sctb &  &  & 692 & 0.31\% \\ \midrule
tha.pdtb.tdtb & Thai & \begin{tabular}[c]{@{}l@{}}Tai-Kadai, \\ Kam-Tai\end{tabular} & 10865 & 4.84\% \\ \midrule
tur.pdtb.tdb & \multirow{2}{*}{Turkish} & \multirow{2}{*}{\begin{tabular}[c]{@{}l@{}}Altaic, \\ Turkic\end{tabular}} & 3185 & 1.42\% \\
tur.pdtb.tedm &  &  & 577 & 0.26\% \\ \midrule
zho.dep.scidtb & \multirow{4}{*}{Mandarin} & \multirow{4}{*}{\begin{tabular}[c]{@{}l@{}}Sino-Tibetan, \\ Chinese\end{tabular}} & 1298 & 0.58\% \\
zho.pdtb.cdtb &  &  & 5270 & 2.35\% \\
zho.rst.gcdt &  &  & 8413 & 3.75\% \\
zho.rst.sctb &  &  & 692 & 0.31\% \\ \bottomrule
\end{tabular}%
}

\caption{Overview of datasets in DISRPT 2023 for the discourse relation classification task.}
\label{tab:disrpt-overview}
\end{table}

\section{Models and Probe Training}
\label{app:models-hyperparams}

\paragraph{LLMs.} \autoref{tab:llms-overview} provides an overview of the examined LLMs in this paper, along with their number of parameters, multilingual capabilities, and the proportion of languages in DISRPT covered by the advertised supported languages.   

\paragraph{Hyperparameters.} We train our probes using AdamW \citep{loshchilovDecoupledWeightDecay2018} with a batch size of $64$, learning rate of $0.0001$, and weight decay of $0.0001$. To mitigate the class imbalance that is inherent to discourse relation data, we use a weighted cross entropy loss, where each class' loss is weighted by the inverse square root of the number of samples in the respective class. For the hidden layer, we choose a dimension of $D$=$512$. For the input as well as the hidden layer, we regularize adding a Dropout of $0.2$. Furthermore, we use layer normalization in the hidden layer. For the probes over all model attention scores, depending on the number of samples contained in the train dataset, we train for $60$ epochs and increase that number to ensure at least $10000$ gradient update steps on smaller datasets. For the layer-wise probes, we reduce the number of epochs to $20$ as the probes converge much faster on these smaller representations. 
Because token-length of encoded sequence quadratically scales the GPU memory required to process the attention matrix, we had to lower the maximum window length for the larger models. Namely, we reduced the maximum window sizes from $N_{max}$=$4000$ to $N_{max}$=$3800$ for \texttt{Aya-Expanse-32B} and to $N_{max}$=$3400$ for \texttt{Aya-23-35B}, \texttt{Llama3-70B}, and \texttt{Qwen2.5-72B}.

\paragraph{Compute.} For each model, we compute one pass over the dataset computing all the attention representations which we cache for the probing experiments. For the smaller models, a cluster equipped with eight Nvidia A100 GPUs was used for around $80$ hours. For the large models of size $70$B and $72$B, we used a cluster equipped with four Nvidia H200 GPUs for about $30$ hours.

\paragraph{Code License.} As specified in the code repository,\!\footnote{\url{https://github.com/mainlp/discourse_probes}} we release our code under MIT license.

\begin{table*}[t]
    \centering
    \resizebox{\textwidth}{!}{%
  \setlength\tabcolsep{3pt}
\begin{tabular}{@{}l|c|c|c|ccccccccccccc|c@{}}
\toprule
\textbf{\begin{tabular}[c]{@{}l@{}}model \\ name\end{tabular}} & \multicolumn{1}{c|}{\textbf{\begin{tabular}[c]{@{}c@{}}model \\ family\end{tabular}}} & \multicolumn{1}{c|}{\textbf{\begin{tabular}[c]{@{}c@{}}\# of \\ params\end{tabular}}} & \multicolumn{1}{c|}{\textbf{\begin{tabular}[c]{@{}c@{}}\# languages\\ supported\end{tabular}}} & \multicolumn{1}{c}{\textbf{deu}} & \multicolumn{1}{c}{\textbf{eng}} & \multicolumn{1}{c}{\textbf{eus}} & \multicolumn{1}{c}{\textbf{fas}} & \multicolumn{1}{c}{\textbf{fra}} & \multicolumn{1}{c}{\textbf{nld}} & \multicolumn{1}{c}{\textbf{por}} & \multicolumn{1}{c}{\textbf{rus}} & \multicolumn{1}{c}{\textbf{spa}} & \multicolumn{1}{c}{\textbf{tur}} & \multicolumn{1}{c}{\textbf{zho}} & \multicolumn{1}{c}{\textbf{tha}} & \multicolumn{1}{c|}{\textbf{ita}} & \multicolumn{1}{c}{\textbf{\begin{tabular}[c]{@{}c@{}}fraction \\ supported\end{tabular}}} \\ \midrule
Qwen2.5-0.5B & \multirow{7}{*}{Qwen 2.5} & 0.49B & \multirow{7}{*}{29} & \cmark & \cmark & \xmark & \cmark & \cmark & \xmark & \cmark & \cmark & \cmark & \xmark & \cmark & \cmark & \cmark & \multirow{7}{*}{0.769} \\
Qwen2.5-1.5B &  & 1.54B &  & \cmark & \cmark & \xmark & \cmark & \cmark & \xmark & \cmark & \cmark & \cmark & \xmark & \cmark & \cmark & \cmark &  \\
Qwen2.5-3B &  & 2.77B &  & \cmark & \cmark & \xmark & \cmark & \cmark & \xmark & \cmark & \cmark & \cmark & \xmark & \cmark & \cmark & \cmark &  \\
Qwen2.5-7B &  & 7.61B &  & \cmark & \cmark & \xmark & \cmark & \cmark & \xmark & \cmark & \cmark & \cmark & \xmark & \cmark & \cmark & \cmark &  \\
Qwen2.5-14B &  & 14.7B &  & \cmark & \cmark & \xmark & \cmark & \cmark & \xmark & \cmark & \cmark & \cmark & \xmark & \cmark & \cmark & \cmark &  \\
Qwen2.5-32B &  & 32.5B &  & \cmark & \cmark & \xmark & \cmark & \cmark & \xmark & \cmark & \cmark & \cmark & \xmark & \cmark & \cmark & \cmark &  \\
Qwen2.5-72B &  & 72.7B &  & \cmark & \cmark & \xmark & \cmark & \cmark & \xmark & \cmark & \cmark & \cmark & \xmark & \cmark & \cmark & \cmark &  \\ \midrule
Llama-3.2-1B & \multirow{4}{*}{Llama 3} & 1.23B & \multirow{4}{*}{8} & \cmark & \cmark & \xmark & \xmark & \cmark & \xmark & \cmark & \xmark & \cmark & \xmark & \xmark & \cmark & \cmark & \multirow{9}{*}{0.538} \\
Llama-3.2-3B &  & 3.21B &  & \cmark & \cmark & \xmark & \xmark & \cmark & \xmark & \cmark & \xmark & \cmark & \xmark & \xmark & \cmark & \cmark &  \\
Llama-3.1-8B &  & 8.03B &  & \cmark & \cmark & \xmark & \xmark & \cmark & \xmark & \cmark & \xmark & \cmark & \xmark & \xmark & \cmark & \cmark &  \\
Llama-3.1-70B &  & 70.6B &  & \cmark & \cmark & \xmark & \xmark & \cmark & \xmark & \cmark & \xmark & \cmark & \xmark & \xmark & \cmark & \cmark &  \\ \cmidrule(r){1-17}
bloom-560m & \multirow{5}{*}{Bloom} & 0.56B & \multirow{5}{*}{46} & \xmark & \cmark & \cmark & \cmark & \cmark & \xmark & \cmark & \xmark & \cmark & \xmark & \cmark & \xmark & \xmark &  \\
bloom-1b1 &  & 1.07B &  & \xmark & \cmark & \cmark & \cmark & \cmark & \xmark & \cmark & \xmark & \cmark & \xmark & \cmark & \xmark & \xmark &  \\
bloom-3b &  & 3B &  & \xmark & \cmark & \cmark & \cmark & \cmark & \xmark & \cmark & \xmark & \cmark & \xmark & \cmark & \xmark & \xmark &  \\
bloom-7b1 &  & 7.07B &  & \xmark & \cmark & \cmark & \cmark & \cmark & \xmark & \cmark & \xmark & \cmark & \xmark & \cmark & \xmark & \xmark &  \\
bloomz-7b1 &  & 7.07B &  & \xmark & \cmark & \cmark & \cmark & \cmark & \xmark & \cmark & \xmark & \cmark & \xmark & \cmark & \xmark & \xmark &  \\ \midrule
aya-expanse-8b & \multirow{4}{*}{Aya} & 8.03B & \multirow{4}{*}{23} & \cmark & \cmark & \xmark & \cmark & \cmark & \cmark & \cmark & \cmark & \cmark & \cmark & \cmark & \xmark & \cmark & \multirow{4}{*}{0.846} \\
aya-expanse-32b &  & 32.3B &  & \cmark & \cmark & \xmark & \cmark & \cmark & \cmark & \cmark & \cmark & \cmark & \cmark & \cmark & \xmark & \cmark &  \\
aya-23-8B &  & 8.03B &  & \cmark & \cmark & \xmark & \cmark & \cmark & \cmark & \cmark & \cmark & \cmark & \cmark & \cmark & \xmark & \cmark &  \\
aya-23-35B &  & 35B &  & \cmark & \cmark & \xmark & \cmark & \cmark & \cmark & \cmark & \cmark & \cmark & \cmark & \cmark & \xmark & \cmark &  \\ \midrule
phi-4 & Phi & 14.7B & 1 & \xmark & \cmark & \xmark & \xmark & \xmark & \xmark & \xmark & \xmark & \xmark & \xmark & \xmark & \xmark & \xmark & 0.077 \\ \midrule
emma-500-llama2-7b & Emma & 6.74B & 546 & \cmark & \cmark & \cmark & \cmark & \cmark & \cmark & \cmark & \cmark & \cmark & \cmark & \cmark & \cmark & \cmark & 1 \\ \midrule
Mistral-Small-24B-Base-2501 & Mistral & 23.6B & 10 & \cmark & \cmark & \xmark & \xmark & \cmark & \xmark & \cmark & \cmark & \cmark & \xmark & \cmark & \xmark & \cmark & 0.615 \\ \bottomrule
\end{tabular}%
}

    \caption{Overview of the included LLMs and their multilingual capabilities for languages in DISRPT 2023.} 
\label{tab:llms-overview}
\end{table*}

\begin{figure}[ht!]
    \vspace{-2pt}
    \centering
    \includegraphics*[trim=0 0 0 400,width=0.9\columnwidth]{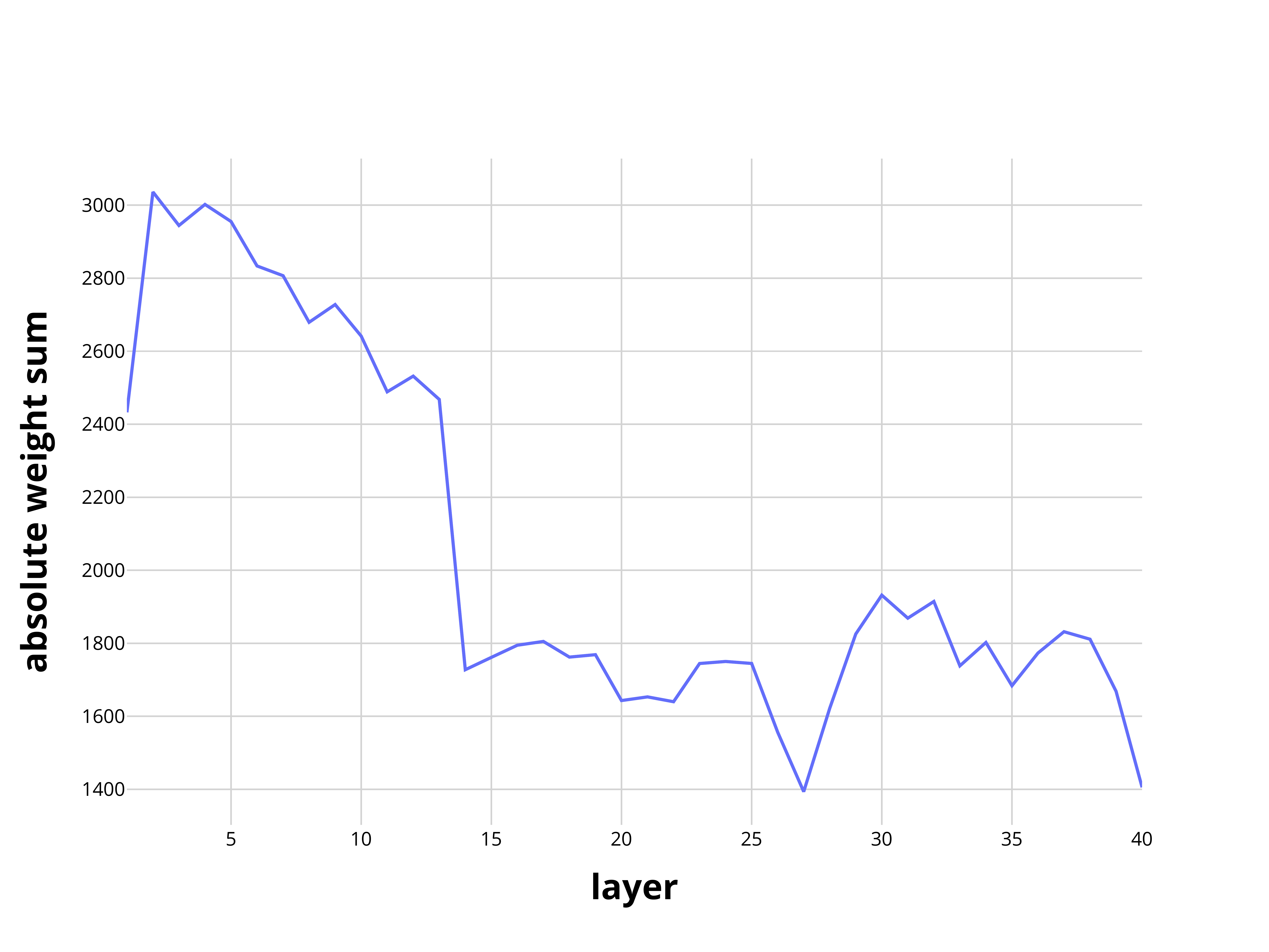}
    \vspace{-8pt}
    \caption{Sum of the absolute dot product between the weight matrices of the probe by layer.}
    \vspace{-10pt}
    \label{fig:layer-wise-weight-sum}
\end{figure}

\section{Additional Results and Analysis}
\label{app:analysis}

\autoref{tab:disrpt-acc-by-model} provides dataset accuracy scores for all examined LLM probes as well as the reference system DisCoDisCo's performance.

\autoref{fig:layer-wise-weight-sum} shows the layer-wise sums of the dot-products of the weight matrices. The magnitude of these scores can be interpreted as a feature importance and confirm that earlier layers play a crucial role in predictions, although the higher layers also show higher scores from layers $30$ to $38$, indicating a progressive refinement of discourse representations.

\begin{table*}[ht]
\centering
\small
\setlength\tabcolsep{1.5pt}

\resizebox{\textwidth}{!}{
\centering

\begin{tabular}{|l|ccccccccccccccccccccccc|cc|}
\toprule
        dataset & \rot{90}{Llama-3.1-70B} & \rot{90}{Llama-3.1-8B} & \rot{90}{Llama-3.2-1B} & \rot{90}{Llama-3.2-3B} & \rot{90}{Mistral-Small-24B-Base-2501} & \rot{90}{Qwen2.5-0.5B} & \rot{90}{Qwen2.5-1.5B} & \rot{90}{Qwen2.5-14B} & \rot{90}{Qwen2.5-32B} & \rot{90}{Qwen2.5-3B} & \rot{90}{Qwen2.5-72B} & \rot{90}{Qwen2.5-7B} & \rot{90}{aya-23-35B} & \rot{90}{aya-23-8B} & \rot{90}{aya-expanse-32b} & \rot{90}{aya-expanse-8b} & \rot{90}{bloom-1b1} & \rot{90}{bloom-3b} & \rot{90}{bloom-560m} & \rot{90}{bloom-7b1} & \rot{90}{bloomz-7b1} & \rot{90}{emma-500-llama2-7b} & \rot{90}{phi-4} & \rot{90}{DisCoDisCo (all)} & \rot{90}{DisCoDisCo} \\ \midrule
        deu.rst.pcc & 41.1 & 35.9 & 31.1 & 33.5 & 39.6 & 24.8 & 24.2 & 36.3 & 37.5 & 26.6 & 41.4 & 30.5 & 43.3 & 39.4 & 38.3 & 38.8 & 23.2 & 27.8 & 19.4 & 22.5 & 23.1 & 36.2 & 35.2 & 25 & 46.5 \\ 
        eng.dep.covdtb & 61.8 & 56.4 & 54.7 & 56.9 & 57.9 & 51.1 & 49.2 & 60.9 & 61.6 & 51.9 & 62.9 & 55.6 & 65.9 & 58.5 & 62.5 & 60.5 & 49.9 & 53.8 & 50.2 & 53.8 & 54.7 & 56.6 & 58.2 & 59.9 & 74.1 \\
        eng.dep.scidtb & 71.8 & 63.2 & 59.5 & 62.7 & 64.0 & 54.4 & 54.4 & 66.2 & 66.8 & 57.8 & 67.3 & 60.5 & 72.4 & 63.9 & 69.5 & 66.3 & 54.5 & 59.1 & 55.0 & 58.8 & 58.9 & 64.2 & 63.9 & 70.7 & 82.7 \\ 
        eng.pdtb.pdtb & 67.4 & 61.6 & 56.7 & 60.6 & 63.8 & 51.3 & 51.7 & 65.6 & 67.1 & 58.5 & 68.4 & 61.5 & 68.8 & 65.3 & 67.9 & 65.3 & 49.4 & 56.2 & 49.4 & 57.1 & 56.6 & 62.0 & 63.4 & 59.1 & 79.9 \\ 
        eng.pdtb.tedm & 48.9 & 53.2 & 49.0 & 47.1 & 52.5 & 43.8 & 45.0 & 52.3 & 55.7 & 49.6 & 53.0 & 52.6 & 51.9 & 58.2 & 54.6 & 56.9 & 40.1 & 46.4 & 45.6 & 47.0 & 46.4 & 53.3 & 52.5 & 40.7 & 58.1 \\ 
        eng.rst.gum & 54.1 & 47.4 & 43.9 & 45.9 & 50.5 & 40.9 & 40.8 & 54.4 & 54.5 & 45.4 & 56.3 & 48.5 & 55.6 & 49.7 & 55.5 & 52.8 & 40.9 & 45.6 & 40.5 & 46.5 & 45.8 & 49.3 & 50.2 & 44.7 & 64 \\
        eng.rst.rstdt & 56.0 & 51.7 & 51.2 & 51.7 & 54.7 & 48.6 & 47.6 & 57.6 & 57.5 & 51.1 & 56.5 & 53.7 & 57.6 & 53.3 & 55.9 & 55.2 & 48.1 & 51.7 & 46.7 & 51.0 & 50.5 & 50.1 & 55.1 & 49.7 & 66.1 \\ 
        eng.sdrt.stac & 48.0 & 44.0 & 41.6 & 43.9 & 44.7 & 40.1 & 41.7 & 46.3 & 46.9 & 42.7 & 48.6 & 42.0 & 49.1 & 45.4 & 47.7 & 45.2 & 39.9 & 41.5 & 38.7 & 41.6 & 41.6 & 43.1 & 45.6 & 45.4 & 60.4 \\ 
        eus.rst.ert & 40.4 & 29.1 & 27.4 & 31.5 & 35.7 & 22.9 & 20.7 & 36.6 & 36.8 & 24.1 & 38.4 & 29.6 & 39.3 & 30.0 & 36.5 & 32.1 & 27.2 & 29.4 & 25.7 & 30.2 & 29.4 & 34.2 & 34.2 & 36 & 62.4 \\ 
        fas.rst.prstc & 49.2 & 43.8 & 43.1 & 46.9 & 47.1 & 36.7 & 39.6 & 49.4 & 49.0 & 43.1 & 51.9 & 46.8 & 49.3 & 48.9 & 49.5 & 48.2 & 37.7 & 40.2 & 33.6 & 40.4 & 38.6 & 47.4 & 46.1 & 46 & 52.7 \\ 
        fra.sdrt.annodis & 48.4 & 43.3 & 41.1 & 42.5 & 45.7 & 41.5 & 41.9 & 53.4 & 52.9 & 40.9 & 51.2 & 47.1 & 50.7 & 44.3 & 47.6 & 47.6 & 37.8 & 42.4 & 39.3 & 43.1 & 42.8 & 43.2 & 49.2 & 26.6 & 56.2 \\ 
        ita.pdtb.luna & 45.7 & 36.1 & 32.7 & 35.4 & 40.2 & 18.2 & 21.1 & 39.8 & 41.7 & 28.7 & 43.7 & 34.6 & 50.9 & 39.8 & 44.5 & 43.9 & 12.9 & 23.4 & 16.6 & 26.7 & 24.2 & 39.9 & 35.0 & 47.1 & 52.1 \\ 
        nld.rst.nldt & 40.0 & 31.9 & 34.9 & 33.5 & 34.5 & 32.8 & 33.5 & 40.7 & 42.4 & 33.4 & 44.9 & 35.4 & 41.3 & 34.9 & 40.6 & 35.0 & 27.4 & 30.1 & 25.0 & 31.3 & 31.5 & 34.5 & 38.0 & 37.2 & 56.6 \\ 
        por.pdtb.crpc & 65.9 & 58.8 & 54.7 & 57.5 & 61.5 & 52.4 & 50.9 & 64.9 & 66.2 & 54.9 & 68.0 & 60.1 & 67.4 & 60.7 & 64.4 & 63.7 & 47.0 & 54.0 & 50.0 & 53.6 & 53.3 & 63.0 & 63.8 & 55.9 & 75.4 \\ 
        por.pdtb.tedm & 57.1 & 51.6 & 43.0 & 47.4 & 49.3 & 39.3 & 41.1 & 55.2 & 57.0 & 45.1 & 58.7 & 49.6 & 61.4 & 53.2 & 57.4 & 55.4 & 39.7 & 43.9 & 39.8 & 46.7 & 45.7 & 50.4 & 54.1 & 49.2 & 66.2 \\ 
        por.rst.cstn & 63.9 & 57.9 & 55.6 & 57.8 & 59.7 & 55.6 & 52.9 & 62.4 & 61.9 & 54.8 & 62.5 & 57.2 & 65.3 & 58.9 & 62.2 & 59.9 & 49.0 & 55.2 & 49.1 & 56.5 & 54.8 & 57.2 & 59.0 & 53.7 & 68.8 \\ 
        rus.rst.rrt & 56.0 & 52.4 & 50.4 & 51.3 & 53.9 & 47.6 & 47.1 & 57.2 & 57.9 & 50.6 & 59.2 & 53.7 & 58.6 & 55.7 & 58.5 & 55.3 & 41.8 & 45.1 & 42.9 & 45.2 & 42.3 & 54.6 & 54.5 & 50.6 & 64.7 \\ 
        spa.rst.rststb & 48.4 & 40.0 & 39.7 & 39.5 & 46.9 & 35.2 & 34.9 & 44.6 & 47.7 & 39.2 & 47.6 & 43.7 & 52.9 & 47.5 & 50.8 & 48.4 & 31.7 & 38.0 & 31.4 & 40.4 & 38.7 & 46.4 & 40.9 & 42 & 61 \\ 
        spa.rst.sctb & 62.9 & 66.9 & 63.5 & 63.5 & 65.0 & 60.4 & 56.2 & 60.9 & 61.9 & 62.0 & 68.3 & 63.1 & 69.7 & 64.2 & 68.9 & 66.3 & 51.2 & 55.7 & 50.6 & 55.5 & 56.0 & 64.3 & 59.6 & 46.5 & 66 \\ 
        tha.pdtb.tdtb & 85.2 & 76.3 & 65.0 & 72.1 & 77.6 & 55.2 & 56.7 & 81.5 & 82.1 & 69.3 & 83.4 & 72.7 & 83.4 & 75.7 & 81.9 & 76.4 & 27.3 & 37.1 & 25.1 & 39.0 & 34.4 & 77.6 & 78.4 & 90.6 & 84.3 \\ 
        tur.pdtb.tdb & 52.8 & 47.7 & 42.8 & 44.5 & 48.3 & 39.2 & 38.0 & 54.2 & 56.5 & 43.6 & 56.5 & 48.5 & 57.9 & 51.7 & 56.3 & 49.4 & 33.2 & 34.8 & 30.5 & 36.1 & 32.6 & 50.8 & 50.2 & 40.3 & 68.7 \\ 
        tur.pdtb.tedm & 48.2 & 43.2 & 39.2 & 43.0 & 48.1 & 32.9 & 32.2 & 49.6 & 50.2 & 35.9 & 52.6 & 43.4 & 49.7 & 44.2 & 50.2 & 46.3 & 17.9 & 24.2 & 20.5 & 23.8 & 23.7 & 48.1 & 42.7 & 27.5 & 53.3 \\ 
        zho.dep.scidtb & 60.7 & 48.3 & 36.3 & 51.6 & 48.4 & 32.8 & 30.8 & 49.8 & 53.9 & 39.4 & 52.8 & 42.0 & 61.6 & 48.0 & 56.8 & 53.8 & 33.9 & 37.4 & 32.9 & 43.3 & 42.0 & 46.7 & 47.3 & 53.5 & 72.1 \\ 
        zho.pdtb.cdtb & 75.9 & 71.2 & 70.6 & 72.6 & 75.4 & 62.7 & 65.4 & 76.3 & 77.1 & 66.2 & 78.0 & 71.8 & 76.3 & 74.5 & 78.4 & 74.3 & 57.2 & 68.7 & 55.8 & 66.3 & 66.8 & 65.6 & 73.7 & 68.3 & 93.8 \\ 
        zho.rst.gcdt & 52.4 & 46.1 & 41.6 & 43.1 & 48.5 & 35.5 & 35.6 & 54.9 & 53.4 & 44.1 & 53.9 & 47.9 & 56.4 & 50.0 & 54.9 & 51.5 & 36.0 & 41.6 & 37.5 & 43.8 & 43.0 & 51.9 & 49.5 & 45.5 & 64.3 \\ 
        zho.rst.sctb & 57.5 & 45.4 & 33.5 & 40.8 & 44.8 & 35.2 & 38.0 & 44.8 & 49.2 & 35.2 & 54.0 & 41.9 & 56.1 & 38.9 & 51.9 & 50.1 & 41.5 & 35.8 & 38.9 & 39.1 & 36.9 & 42.5 & 40.6 & 34 & 56.6 \\ 
        \midrule
        average & 56.1 & 50.1 & 46.3 & 49.1 & 52.2 & 42.0 & 42.0 & 54.5 & 55.6 & 45.9 & 56.9 & 49.8 & 58.2 & 52.1 & 56.3 & 53.8 & 38.3 & 43.0 & 38.1 & 43.8 & 42.9 & 51.3 & 51.6 & 47.9 & 65.7 \\ 
\bottomrule
\end{tabular}
}

\caption{\textbf{Dataset accuracy scores by LLM probe averaged over five runs.} The last two columns refer to the DisCoDisCo reference system trained on all languages (all) and trained a multiple models with different encoders per dataset.}
\label{tab:disrpt-acc-by-model}
\end{table*}

\begin{table*}[ht]
\centering
\small
\setlength\tabcolsep{1.5pt}
\resizebox{\textwidth}{!}{
\begin{tabular}{|l|ccccccccccccccccccccccc|}
\toprule
          & \rot{90}{Llama-3.1-70B} & \rot{90}{Llama-3.1-8B} & \rot{90}{Llama-3.2-1B} & \rot{90}{Llama-3.2-3B} & \rot{90}{Mistral-Small-24B-Base-2501} & \rot{90}{Qwen2.5-0.5B} & \rot{90}{Qwen2.5-1.5B} & \rot{90}{Qwen2.5-14B} & \rot{90}{Qwen2.5-32B} & \rot{90}{Qwen2.5-3B} & \rot{90}{Qwen2.5-72B} & \rot{90}{Qwen2.5-7B} & \rot{90}{aya-23-35B} & \rot{90}{aya-23-8B} & \rot{90}{aya-expanse-32b} & \rot{90}{aya-expanse-8b} & \rot{90}{bloom-1b1} & \rot{90}{bloom-3b} & \rot{90}{bloom-560m} & \rot{90}{bloom-7b1} & \rot{90}{bloomz-7b1} & \rot{90}{emma-500-llama2-7b} & \rot{90}{phi-4} \\

\midrule

$\mathbf{C}$ (inter) & 53.8 & 45.5 & 41.5 & 43.4 & 45.5 & 35.1 & 35.6 & 49.1 & 52.9 & 40.8 & 55.8 & 42.4 & 53.7 & 47.6 & 54.3 & 47.0 & 32.4 & 39.9 & 32.3 & 40.3 & 40.1 & 44.4 & 46.9 \\
$\mathbf{D}_1, \mathbf{D}_2$ (intra) & 46.0 & 38.2 & 35.2 & 37.3 & 40.1 & 33.3 & 32.9 & 43.5 & 44.5 & 33.4 & 47.4 & 38.3 & 45.5 & 36.0 & 39.9 & 40.3 & 28.5 & 30.8 & 29.0 & 31.9 & 31.3 & 40.4 & 41.3 \\
$\mathbf{D}_1, \mathbf{D}_2, \mathbf{C}$ (all) & \underline{55.9} & \underline{50.8} & \underline{46.2} & \underline{48.6} & \underline{52.5} & \underline{41.6} & \underline{42.2} & \underline{54.2} & \underline{55.8} & \underline{46.0} & \underline{56.8} & \underline{50.3} & \underline{58.2} & \underline{52.6} & \underline{56.0} & \underline{54.0} & \underline{37.8} & \underline{42.6} & \underline{37.0} & \underline{43.5} & \underline{43.9} & \underline{51.4} & \underline{51.9} \\
\bottomrule

\end{tabular}}

\caption{\textbf{Overall accuracy scores by LLM probe averaged over five runs training on the entire DISRPT (probe representation ablation).} We ablate different types of attention representations used for probing and find that using all described matrices $\mathbf{D}_1, \mathbf{D}_2, \mathbf{C}$ yields the best results.}
\label{tab:disrpt-acc-by-model-ablated}
\end{table*}

\begin{table*}[ht]
\centering
\small
\setlength\tabcolsep{1.5pt}
\resizebox{\textwidth}{!}{
\begin{tabular}{|l|ccccccccccccccccccccccc|}
\toprule
          & \rot{90}{Llama-3.1-70B} & \rot{90}{Llama-3.1-8B} & \rot{90}{Llama-3.2-1B} & \rot{90}{Llama-3.2-3B} & \rot{90}{Mistral-Small-24B-Base-2501} & \rot{90}{Qwen2.5-0.5B} & \rot{90}{Qwen2.5-1.5B} & \rot{90}{Qwen2.5-14B} & \rot{90}{Qwen2.5-32B} & \rot{90}{Qwen2.5-3B} & \rot{90}{Qwen2.5-72B} & \rot{90}{Qwen2.5-7B} & \rot{90}{aya-23-35B} & \rot{90}{aya-23-8B} & \rot{90}{aya-expanse-32b} & \rot{90}{aya-expanse-8b} & \rot{90}{bloom-1b1} & \rot{90}{bloom-3b} & \rot{90}{bloom-560m} & \rot{90}{bloom-7b1} & \rot{90}{bloomz-7b1} & \rot{90}{emma-500-llama2-7b} & \rot{90}{phi-4} \\

\midrule

mean & 55.6 & 48.7 & 45.2 & 47.8 & 48.6 & 39.4 & 40.5 & 50.5 & 51.7 & 42.9 & 54.4 & 46.4 & 54.2 & 48.6 & 53.1 & 50.1 & 35.1 & 40.0 & 34.6 & 40.1 & 39.9 & 49.7 & 48.3 \\
max & 55.9 & 50.8 & 46.2 & \underline{48.6} & 52.5 & 41.6 & 42.2 & \underline{54.2} & 55.8 & 46.0 & \underline{56.8} & 50.3 & 58.2 & 52.6 & 56.0 & 54.0 & 37.8 & 42.6 & 37.0 & \underline{43.5} & 43.9 & 51.4 & 51.9 \\
mean,max & \underline{56.1} & \underline{51.0} & \underline{46.5} & \underline{48.6} & \underline{52.8} & \underline{42.6} & \underline{42.8} & 54.0 & \underline{56.0} & \underline{47.1} & 56.1 & \underline{50.8} & \underline{58.4} & \underline{52.8} & \underline{56.9} & \underline{54.2} & \underline{38.5} & \underline{43.7} & \underline{38.4} & 43.2 & \underline{44.0} & \underline{51.5} & \underline{52.0} \\

\bottomrule

\end{tabular}}

\caption{\textbf{Overall accuracy scores by LLM probe averaged over five runs training on the entire DISRPT (pooling strategy ablation).} We ablate different attention-head-wise pooling strategies, namely mean pooling, maximum pooling, and the concatenation of both on the attention score matrices $\mathbf{D}_1, \mathbf{D}_2, \mathbf{C}$. We find that concatenating both generally yields the best results, though usually only by a margin of less than $1\%$. To keep the size of the representations shorter, we thus opt to only keep the maximum pooling.}
\label{tab:disrpt-acc-by-model-ablated2}
\end{table*}

\begin{figure*}[!ht]
    \centering
    \subfloat[\textsc{thematic}]{\includegraphics*[width=0.5\textwidth]{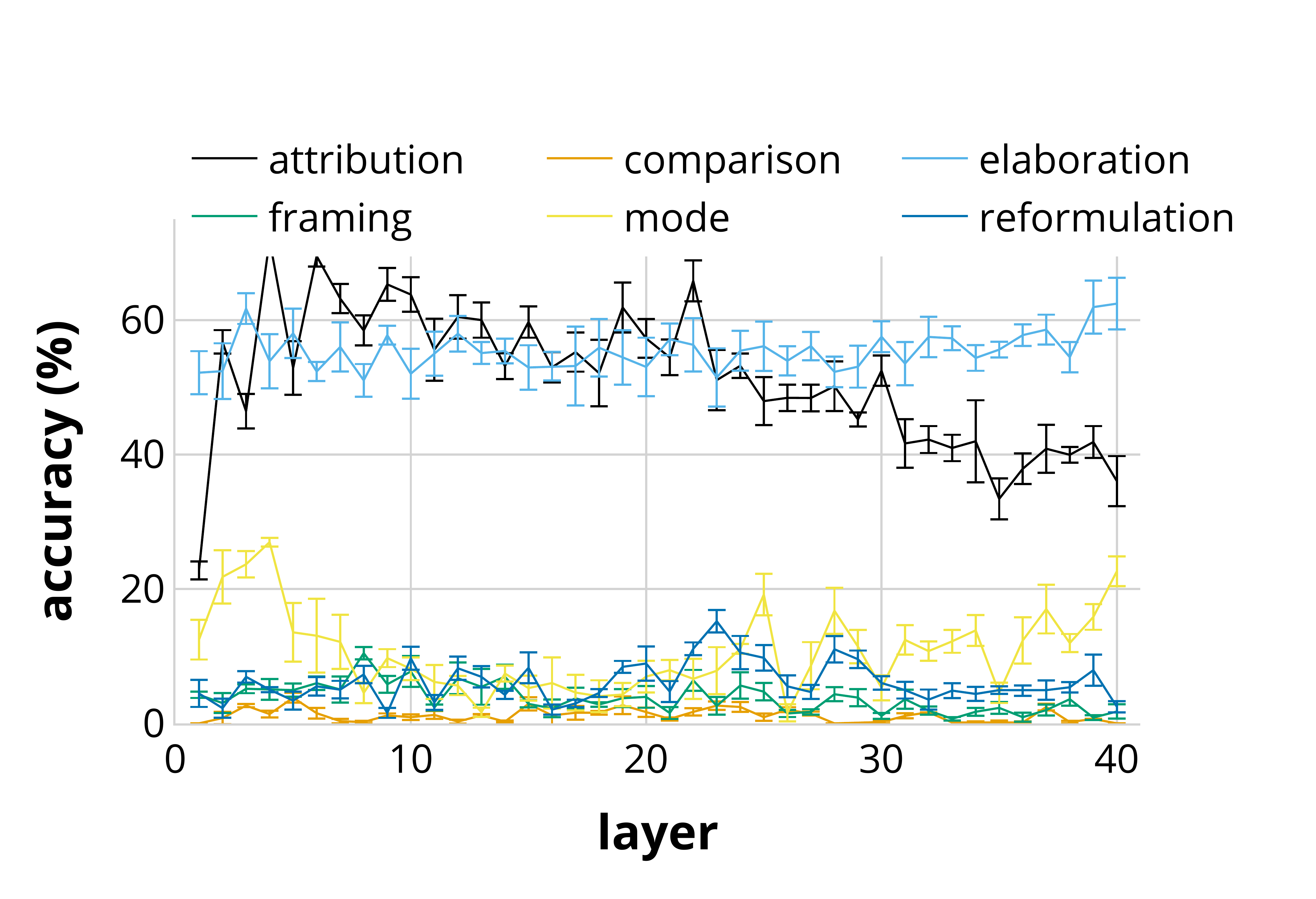}\label{fig:layer-wise-lab-acc-thematic}}
    \subfloat[\textsc{topic-management}]{\includegraphics*[width=0.5\textwidth]{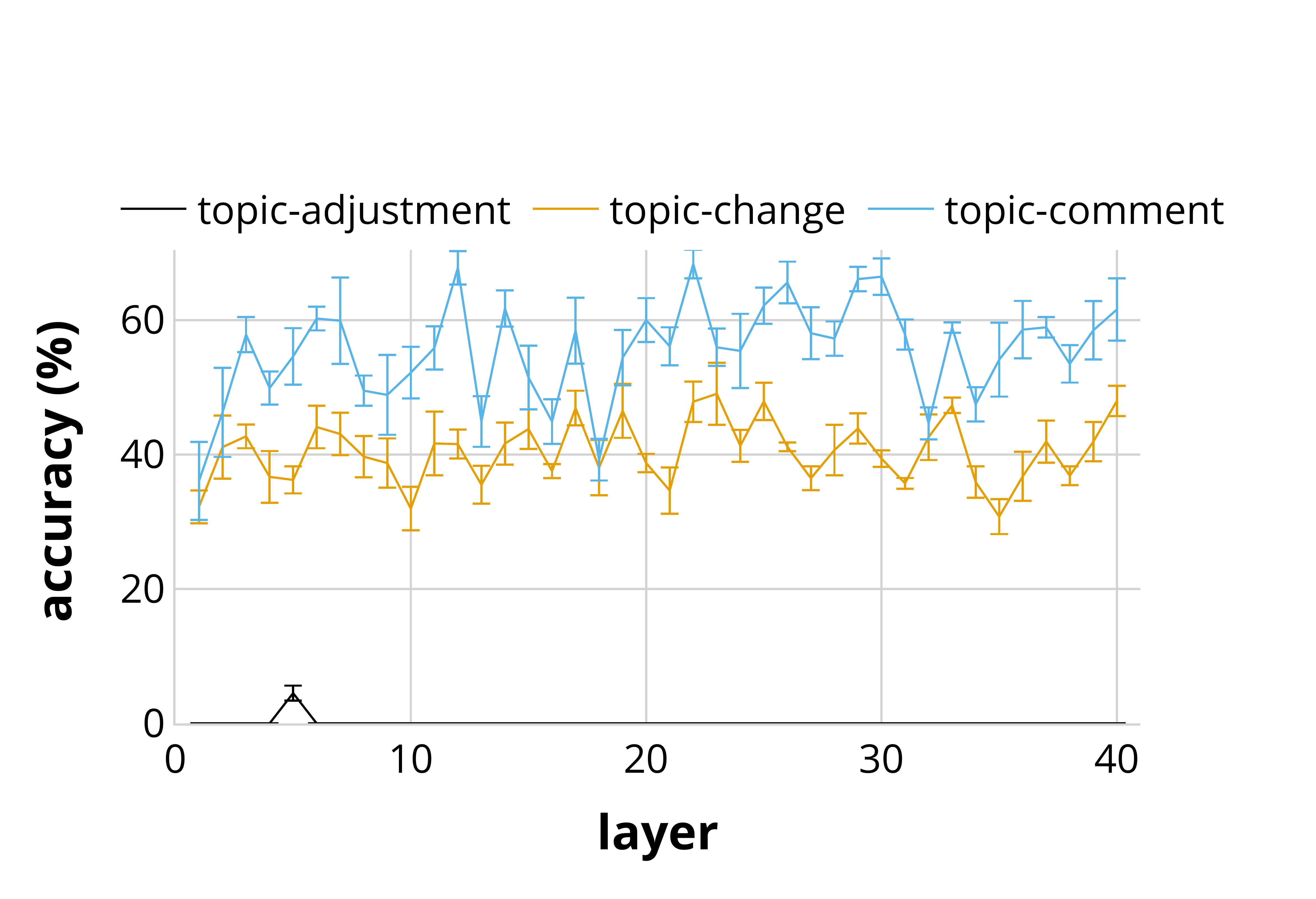}\label{fig:layer-wise-acc-topic}}
    \vspace{-2pt}
    \caption{\textbf{Layer-wise probe performance by relation classes.} Mean accuracy over five runs.} 
    \vspace{-8pt}
    \label{fig:layer-wise-acc-rel-additional}
\end{figure*}

\section{Use of AI Assistants}
\label{app:ai-use-statement}

The implementation of this work has been written with the support of code completions of an AI coding assistant, namely GitHub Copilot. Completions were mostly single lines up to a few lines of code and were always checked carefully to ensure their functionality and safety. Furthermore, we did our best to avoid accepting code completions that would be incompatible with the license of our code or could be regarded as plagiarism. We also include this statement in the \texttt{README.md} of the codebase.

\end{document}